# Thinking Outside the Bots:

# Human Ideas in AI's Semantic Blind Spots Are Not Inferior

Kibum Moon[1]*, Dan R. Johnson[2]*, Kostadin Kushlev[1], Benjamin Lira[3], Indre Viskontas[4],

James C. Kaufman[5], Roger E. Beaty[6], Angela L. Duckworth[7], and Adam E. Green[1]

*These authors contributed equally to this work.

*This manuscript has not been peer-reviewed.*

[1]Department of Psychology, Georgetown University

[2]Department of Cognitive and Behavioral Science, Washington and Lee University

[3]The Wharton School, University of Pennsylvania

[4]Department of Psychology, University of San Francisco

[5]Neag School of Education, University of Connecticut

[6]Department of Psychology, Pennsylvania State University

[7]Department of Psychology, University of Pennsylvania

**Abstract**

AI-generated outputs are frequently perceived as highly creative, but also tend to be more similar to each other (homogenized) than human ideas. Does the human heterogeneity that extends beyond AI reflect valuable thinking that AI misses? Alternatively, is greater human heterogeneity driven by lower-quality ideas and less effective thinking that AI avoids? We developed Expansion of Divergent Generative Expression (EDGE) to map AI's semantic "blind spots" by assessing how strongly/weakly individual human ideas were represented in a task-matched AI semantic comparison set. Specifically, we compared human-generated vs. AI-generated ideas across three open-ended tasks: college admissions essays, alternative uses for everyday objects, and interpersonal advice. Across all three tasks, human responses extended into semantic regions sparsely covered by AI outputs. These distinctively human responses showed no meaningful deficits in creativity, quality, or effectiveness, were associated with equal-or-better academic performance, and with credibly superior ratings and achievement for some outcomes. These ideas were thus not merely poor-performing variation; they performed as well as or better than ideas that overlapped more strongly with AI ideas. More distinctive human ideas tended to receive higher creativity ratings, but this relationship was weak or reversed for AI ideas. Taken together, these results suggest that differentiation among human-generated ideas does not reflect poor thinking that AI can safely omit, and instead may afford both individual and population-level advantages.

**Thinking Outside the Bots:**

**Human Ideas in AI's Semantic Blind Spots Are Not Inferior**

Across creative writing, ideation, and brainstorming, LLM-generated or LLM-assisted responses are often more similar to one another and cover a narrower range of ideas than comparable human responses (Anderson et al., 2024; Doshi & Hauser, 2024; Meincke et al., 2025; Moon et al., 2025, 2026; Padmakumar & He, 2024; Sourati et al., 2026). This pattern is not limited to a single model: Different LLMs often converge on similar responses to the same open-ended prompts (Jiang et al., 2025; Wenger & Kenett, 2025), and convergence appears to be increasing among more recently released models (Patel et al., 2026). Researchers have therefore described reduced diversity as a risk or loss for individual and collective creativity (Doshi & Hauser, 2024; Moon et al., 2025; Padmakumar & He, 2024; Sourati et al., 2026). However, the question of whether and to what extent AI poses this risk to creativity appears to hinge on resolving two competing alternatives: does AI homogenization relative to human diversity reflect limitations on the range of good ideas or a reduction in the generation of bad ideas?

One possibility is that AI homogenization underrepresents valuable human ideas. By learning to predict the next token from statistical distributions in language, LLMs may favor common linguistic and conceptual patterns, leaving less typical ideas underrepresented (Resnik, 2025; Sourati et al., 2026). Imbalances in web-based training data can privilege some perspectives over others (Bender et al., 2021), while supervised fine-tuning and preference-based alignment can further concentrate generation around a narrower set of responses (Kirk et al., 2024; Klypa & Cherednichenko, 2026; Murthy et al., 2025). Such convergence may already be present before alignment, with evidence from base models suggesting that pretraining may contribute to output homogeneity (Fortier et al., 2026). On this account, LLMs may leave out an idea because it is uncommon, not because it lacks value.

A second possibility is that human ideas are more heterogeneous than AI-generated ideas because humans generate many low-value ideas that AI avoids. Human variation can include noise, idiosyncratic responses, or weak fits to task demands. Consistent with this possibility, human-only responses can be collectively more diverse than AI-assisted responses yet receive lower ratings on measures of creativity or writing quality (Doshi & Hauser, 2024; Holzner et al., 2025; Luchini, Pronchick, et al., 2025; Meincke et al., 2025). These comparisons do not establish, however, whether the human responses that LLMs leave weakly represented are themselves of lower quality.

Distinguishing between these possibilities requires examining whether human responses weakly represented in LLM outputs nevertheless retain value. Prior studies have documented reduced collective diversity by comparing collections of human, AI-generated, or AI-assisted responses using aggregate measures of semantic diversity (Anderson et al., 2024; Patel et al., 2026; Wenger & Kenett, 2025) and variation in linguistic features (Alvero et al., 2024). Even measures that capture individual contributions to collective diversity (Luchini, Pronchick, et al., 2025) do not, by themselves, establish whether human responses weakly represented in LLM outputs retain value. Prior research has therefore demonstrated reduced diversity, but not whether the ideas beyond the narrower range of LLM outputs represent a meaningful loss. Addressing this gap requires identifying human responses that LLMs weakly represent and evaluating their task-relevant outcomes independently of the models. We call the regions occupied by these responses *AI semantic blind spots*: regions of task-relevant semantic possibility that fall outside the effective reach of LLM generation.

To address this question, we introduce Expansion of Divergent Generative Expression (EDGE), a measure designed to identify responses that occupy less well-represented parts of a semantic space. EDGE gives each response a score based on how isolated it is from dense groups of responses and how weakly it aligns with any single group. Higher EDGE scores indicate that similar responses are relatively uncommon in the comparison set. Using EDGE, we addressed two primary questions. First, were human responses more likely than matched LLM responses to occupy semantically sparse regions of their shared response space? Second, among human responses, was weaker representation by matched LLM outputs associated with meaningfully poorer task-relevant outcomes? If AI semantic blind spots consist primarily of noise, idiosyncratic variation, or poor fits to task demands, then human responses weakly represented by matched LLM outputs should perform meaningfully worse than responses more strongly represented by those outputs. Instead, we predicted that human responses in AI blind spots would not be inferior, i.e., that their outcomes would be as good as or better than outcomes for ideas that are more similar to AI.

# Method

## Study Overview

We analyzed matched human and LLM-generated responses across three open-ended generative tasks: college admissions essays, alternative uses for everyday objects, and advice on an interpersonal problem. These tasks span consequential writing, brief idea generation, and everyday problem solving, allowing us to address our research questions across diverse forms of idea generation. The analyses asked whether matched LLM responses left task-specific semantic blind spots occupied by heterogeneous human responses and whether, among human responses, occupying those blind spots came without a substantively meaningful cost to task-relevant outcomes. We addressed the first question through a Coverage Analysis, which scored human and LLM responses together in repeated task-matched samples and compared their mean EDGE scores. We addressed the second question through an Outcome Noninferiority Analysis, which examined how task-relevant outcomes changed as human responses became less well represented by matched LLM outputs and whether any decline was large enough to constitute a substantively meaningful cost.

## Datasets and Response Sources

### *Admissions Essays*

We analyzed 1,600 admissions essays written by college applicants from three U.S. institutions: an elite private university, a regional private university, and a selective liberal-arts college. We focused on essays written by matriculated students so that the essays represented successful applications at each institution and could be linked to post-enrollment academic performance. To minimize the possibility that the human-authored essays were influenced by generative AI assistance, we restricted the sample to application cohorts from 2022 or earlier, before ChatGPT became publicly available.

Admissions essays were written in response to specific prompts, which substantially shaped their content. Because institution and prompt were not fully crossed, each observed institution–prompt combination constituted a distinct matched response context. We therefore treated each combination as a sampling stratum and retained eight strata containing at least 200 essays from the 2018–2022 matriculated sample: one at the elite private university, four at the regional private university, and three at the selective liberal-arts college. To balance sample sizes across institution-prompt strata and limit the computational demands of EDGE estimation, we randomly sampled 200 essays without replacement from each stratum, yielding a human sample of 1,600 essays.

Three prompts were shared by the regional private university and the selective liberal-arts college, so the eight essay sampling strata represented five distinct prompts. For each of these five prompts, GPT-5-mini generated 200 essay responses, yielding a matched AI response pool of 1,000 essays. Each human or AI essay was treated as one response-level unit. See Supplemental Method, “Admissions Essays,” and Tables S1a–S1b for further details.

### ***Alternative Uses Task***

The Alternative Uses Task (AUT) is a widely used measure of divergent thinking in which participants generate unusual uses for everyday objects. Performance on the AUT task has been shown to be associated with creative activities and achievements in everyday life, including in the arts and sciences (Beaty et al., 2018). We analyzed AUT responses from Patterson et al. (2023) for two object prompts: brick and knife. Human responses from 241 participants yielded 481 participant–object bundles: 240 brick bundles ($n$ = 1,739 individual uses) and 241 knife bundles ($n$ = 1,686 individual uses). Each bundle preserved the participant’s set of responses for a given object as the analysis unit, rather than treating individual uses as independent observations. Of the 241 participants, 240 contributed bundles for both objects, whereas one contributed only a knife bundle. The matched AI response pool comprised 481 GPT-5-mini pseudo-participant bundles generated for the same brick and knife prompts, with one AI bundle corresponding to each human bundle and containing the same number of individual uses. See Supplemental Method, “Alternative Uses Task,” for generation and matching procedures.

### *Creative Problem-Solving Task*

For the Creative Problem-Solving Task (CPS), we used the Becky's Problem responses from the CPS data reported by Luchini, Maliakkal, et al. (2025). In the CPS, participants read an ill-defined, everyday problem scenario centered on a named character and generate a written solution. Specifically, Becky's Problem asks participants to advise a college student whose roommate is stealing food from the pizzeria where they both work. She is hesitant to tell their employer because her roommate is a close friend and she depends on his income to cover his share of their rent (Reiter-Palmon et al., 1998). The human sample consisted of 1,018 undergraduate-written responses across the Becky datasets, and the AI reference set consisted of 1,018 GPT-5-mini responses generated for the same Becky's Problem prompt. To approximate the mean length of the human responses (53.77 words and 2.58 sentences), we instructed the model to aim for approximately 54 words in about three sentences. Each participant contributed one complete response, and each complete human or AI solution was treated as one response-level unit. See Supplemental Method, "Creative Problem-Solving Task," for the scenario and generation instructions.

## Outcome Measures

External person-level outcomes for admissions essay authors included SAT and cumulative college grade point average (GPA), obtained from university records. ACT composite scores were converted to SAT equivalents using the official 2018 ACT/SAT concordance (ACT, Inc. & College Board, 2018). When both SAT and ACT scores were available, we retained the higher SAT-scale score.

Human-rated response outcomes were available only for human-authored AUT and CPS responses. For the AUT, the original dataset provided an originality score for each use, calculated as the mean of three human ratings on a 1–5 scale (Patterson et al., 2023). For the present analyses, we summarized originality at the participant–object level by averaging these use-level scores. For the CPS, the original dataset provided originality and quality scores for each solution, based on ratings from separate panels of two to four trained raters using 5-point scales (Luchini, Maliakkal, et al., 2025).

To enable direct comparisons between response sources, we applied the same automatic assessment within each task to both human and matched AI-generated responses. Admissions essays were scored using a GPT-4.1-mini evaluator fine-tuned to predict creativity judgments provided by 22 creativity researchers (Moon et al., 2026). The evaluator was developed from ratings of 370 pre-ChatGPT admissions essays, each evaluated by at least three experts. It produced percentile-scaled creativity scores ranging from 0 to 1 and achieved a correlation of .82 with mean expert ratings in a held-out set of 111 essays. AUT and CPS responses were scored using Penn State's Creativity Assessment Platform (CAP; Patterson et al., 2025), which provides language models trained to approximate human creativity judgments. For the AUT, Cross-Lingual Alternate Uses Scoring (CLAUS), an XLM-RoBERTa model trained to predict human originality ratings, scored each individual use. We averaged these predictions across uses within each participant–object bundle. For the CPS, CPS-AI was developed using human ratings of solutions produced by 3,243 participants across 10 studies and was validated for both originality and solution quality (i.e., effectiveness), including on previously unseen problem prompts (Luchini, Maliakkal, et al., 2025). We used CPS-AI to evaluate the originality and effectiveness of each complete CPS response. Additional scoring details are provided in Supplemental Method, "Outcome Measures."

**Ethical Compliance**

The analysis of admissions essays and associated academic records was approved by the Institutional Review Board of [redacted for peer review] (protocol no. [*redacted for peer review*]). The AUT and CPS analyses used existing datasets reported by Patterson et al. (2023) and Luchini, Maliakkal, et al. (2025), respectively.

**EDGE Measure**

EDGE quantifies how weakly a response is represented within a specified, task-matched comparison set. A response receives a higher EDGE score when it lies apart from dense concentrations of comparable responses (i.e., outlyingness), does not align clearly with any single shared semantic pattern (i.e., cluster-membership ambiguity), or both. EDGE is therefore relative to the responses included in the comparison set. Because its calculation uses only their semantic positions, it does not directly measure originality, creativity, quality, or effectiveness; whether EDGE is associated with these attributes is an empirical question.

As outlined in Figure 1, we first represented responses using three open-source sentence-embedding models: NoInstruct-small (Solatorio, 2024b), Sentence-Transformers MPNet (Reimers & Gurevych, 2019; Song et al., 2020), and GIST-Embedding-v0 (Solatorio, 2024a). Using three embedding models reduced dependence on any single representation of semantic similarity. Admissions essays and CPS responses were segmented into sentences, and each sentence was embedded separately. Individual AUT uses were likewise embedded separately. EDGE components were computed across the resulting task-specific embedding rows and then averaged within each analytic unit: across sentences for each admissions essay and CPS response, and across uses for each AUT participant–object bundle.

Within each embedding space, EDGE combined two unsupervised components. First, the Global-Local Outlier Score from Hierarchies (GLOSH), derived from HDBSCAN's hierarchical density structure, measured how far a response lay from local and global concentrations of responses (Campello et al., 2015). Second, fuzzy c-means cluster-membership ambiguity measured how weakly a response aligned with any single semantic cluster (Bezdek et al., 1984). Membership ambiguity was calculated as one minus the response's highest cluster-membership value. The two components were computed separately in each of the three embedding spaces, producing six component scores for every analytic unit. These components were standardized within the analysis-specific scoring set and averaged to produce the final EDGE score. Higher scores indicate weaker representation in the corresponding comparison set through greater outlyingness, greater cluster-membership ambiguity, or both. Figure 2 shows what the two components look like in a two-dimensional display space.

**Figure 1**

*Overview of the EDGE Computation Procedure*

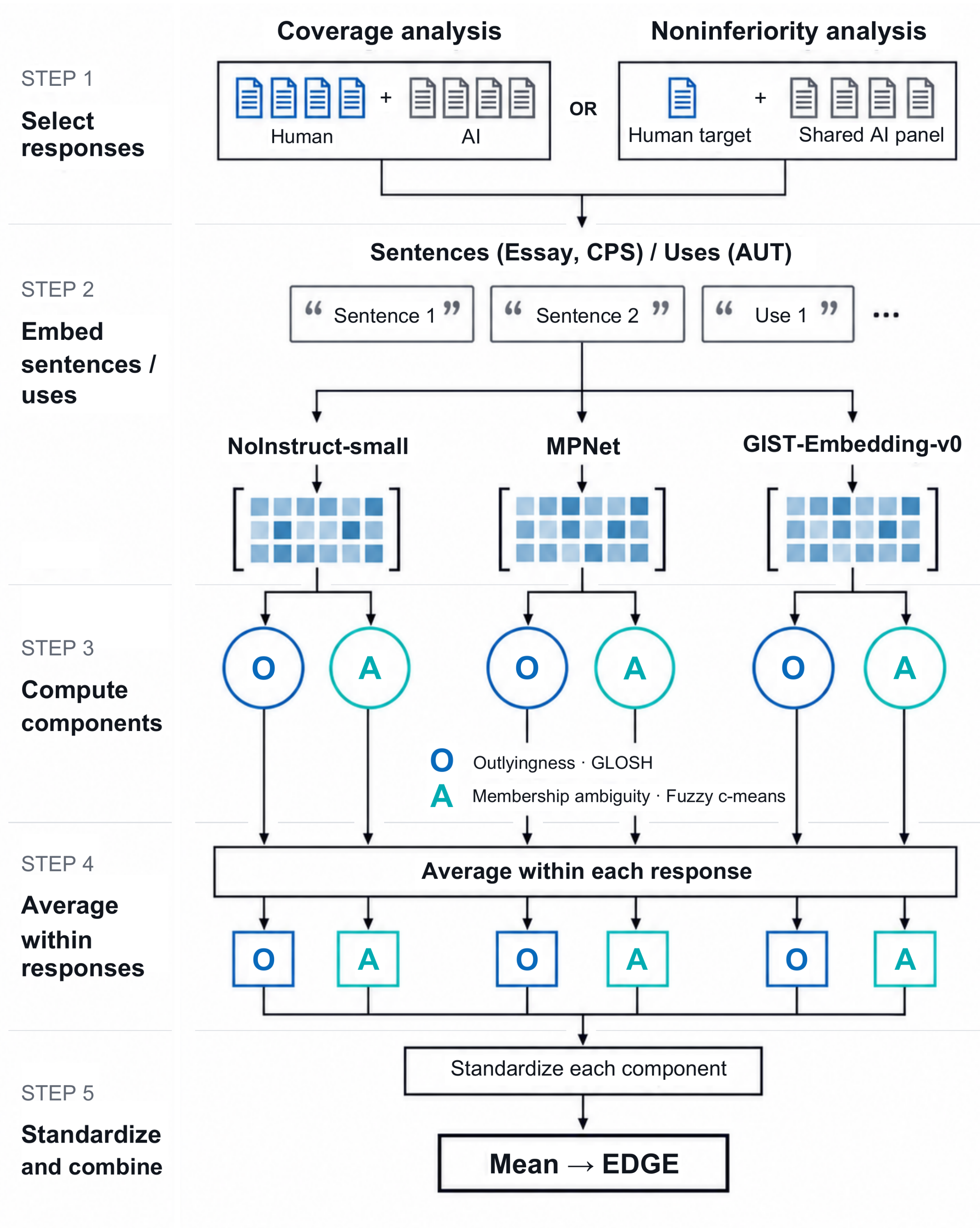

**Figure 2**

*Visual Explanation of the Two EDGE Components*

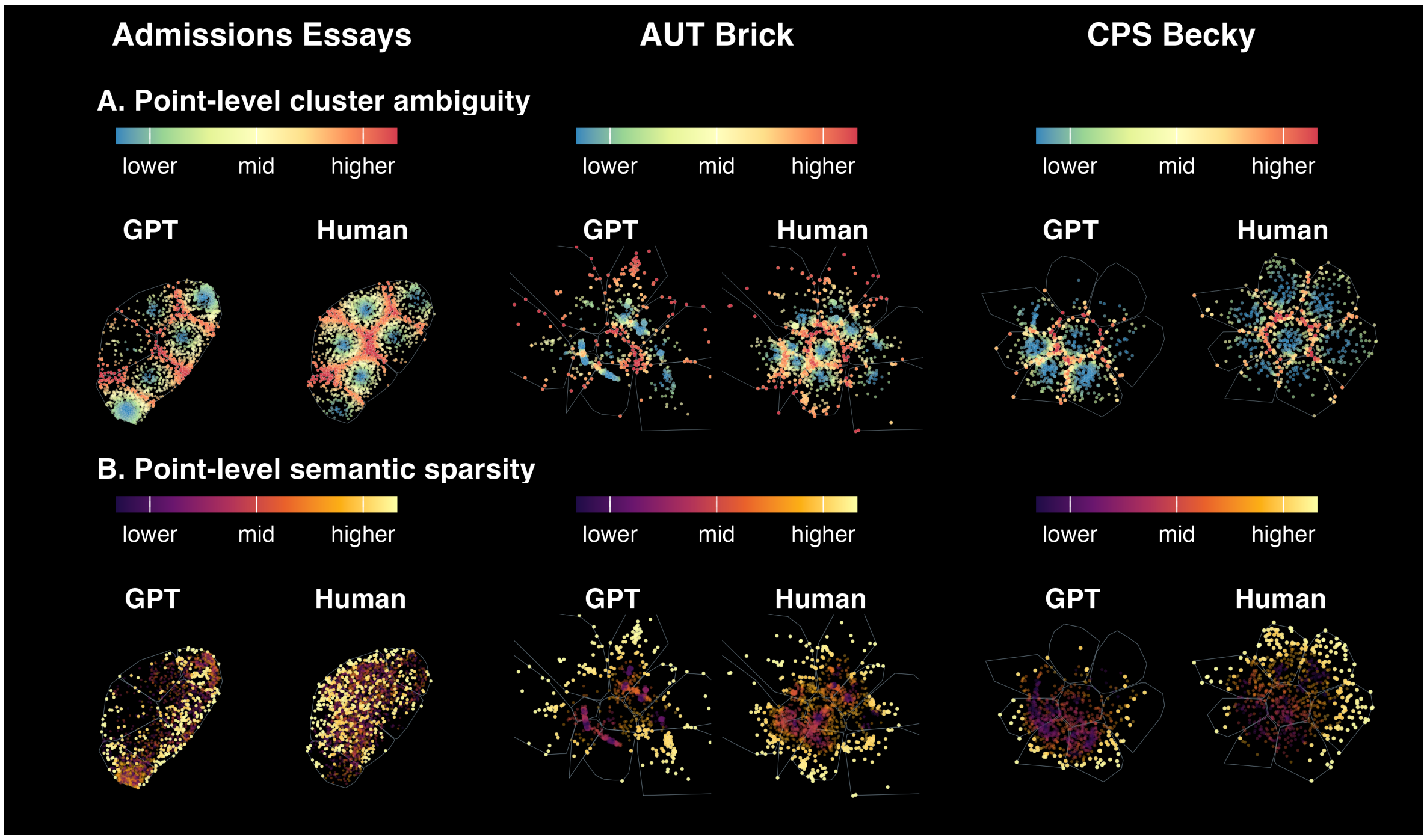

*Note.* Each point represents a sentence for Admissions Essays, an individual use for AUT Brick, or a sentence for CPS Becky. Equal numbers of GPT and human response units were selected within each task. Because these response units could contain different numbers of sentences or uses, the GPT and human panels could contain different numbers of points. Row A visualizes fuzzy c-means cluster membership ambiguity. Row B visualizes semantic sparsity, which serves as a display space analogue of the hierarchical outlyingness captured by GLOSH. Colors indicate displayed values, as shown by the color scales, and larger points in Row B mark the pooled top 15% within each task. The two-dimensional coordinates and visual encodings were used only for visualization. The reported analyses used the original high-dimensional embeddings and EDGE scores aggregated to the response level. Implementation details are provided in Supplemental Method, “EDGE Computation.”

## Analytic Strategy

We conducted two primary analyses followed by an analysis designed to clarify what EDGE captures in human and AI responses. Because EDGE is comparison-dependent, each analysis used the comparison set appropriate to its question. The Coverage Analysis compared human and AI responses within a shared semantic space. The Outcome Noninferiority Analysis instead calculated EDGE separately for each human response using a comparison set comprising that response and a task-matched AI reference set, thereby defining weak representation specifically relative to matched AI outputs.

### *Coverage Analysis*

The Coverage Analysis tested whether human responses were more likely than matched AI responses to occupy sparsely represented regions of their shared semantic space. Human and AI responses were scored together within repeated task-matched draws. Each draw sampled 120 human response units and 120 matched AI response units without replacement. Sentences and individual uses retained the same positions in each model's embedding space across draws, but the clustering structure and EDGE scores were recomputed for each sampled set. This allowed us to assess how the human–AI contrast varied with the composition of the comparison set. We defined the coverage contrast as the mean EDGE score among human responses minus the mean EDGE score among AI responses within the same draw. Positive values indicated that human responses, on average, occupied more sparsely represented regions of the shared response space.

We generated 1,000 draws for each retained essay sampling stratum, AUT object, and CPS scenario. Within each draw index, task-level contrasts were calculated by equally averaging the eight institution–prompt contrasts for admissions essays and the two object-specific contrasts for the AUT; CPS required no aggregation across groups. The distribution of draw-level contrasts represented variation across sampled comparison sets. For each task, we summarized this distribution using its median and the shortest empirical interval containing 95% of the draw-level contrasts (95% resampling interval). A 95% resampling interval entirely above zero supported the prediction that human responses had higher mean EDGE than matched AI responses.

### ***Bayesian Outcome Noninferiority Analysis***

The Bayesian Outcome Noninferiority Analysis tested whether human responses less well represented by matched AI outputs showed meaningfully poorer task-relevant outcomes. In the pooled comparison sets used for the Coverage Analysis, other human responses could make an idea appear well represented even when similar AI responses were rare. We therefore scored each human response separately with a panel of matched AI responses, excluding all other human responses. For each calculation, we combined the target response's sentences or uses with those of the AI panel and refitted the clustering. We refer to the resulting scores as AI-reference EDGE.

For admissions essays, we sampled 120 AI responses without replacement separately for each prompt. For the AUT, we sampled 120 AI responses separately for each object, and for the CPS, we sampled 120 AI responses to Becky's Problem. Each sampled set was held fixed and used to recalculate EDGE for all human responses matched to it. For each reference set, EDGE scores were standardized within the corresponding essay sampling stratum, AUT object, or CPS problem. Outcomes were standardized once within the same matched response groups. We repeated this procedure 10 times, drawing a new AI reference set each time to account for variation in reference-set composition.

Across matched response groups, the estimated consistency of the 10-panel mean ranged from .975 to .999 (Table S6; Supplemental Results, “Stability Across AI-Reference Panels”).

To incorporate variation across the 10 reference sets, we summarized each human response’s standardized EDGE scores by their mean and the standard error of that mean. We then entered the mean EDGE score and its standard error as a measurement-error predictor in Bayesian regression models estimated with *brms* (Bürkner, 2017). This approach incorporated uncertainty arising from reference-set composition while retaining one task-relevant outcome per human response. The resulting EDGE coefficient represented the expected change in the outcome, in standard-deviation units, associated with a one-standard-deviation increase in AI-reference EDGE. We estimated a separate Bayesian measurement-error model for each outcome. Each target’s EDGE score was represented as a latent predictor informed by its mean and standard error across the 10 AI reference sets, using a centered parameterization.

When an analysis included multiple matched response groups, the model allowed the mean of latent EDGE to vary across groups and estimated a separate outcome intercept and EDGE slope for each group. The matched response groups were the eight institution–prompt sampling strata for admissions essays and the brick and knife object groups for the AUT. In each posterior draw, the task-level coefficient was defined as the arithmetic mean of the eligible group-specific slopes. Because all but one AUT participant contributed responses to both objects, the AUT outcome models additionally included a participant random intercept. CPS involved a single problem and therefore required one latent-EDGE intercept, one outcome intercept, and one EDGE slope. All models used the default *brms* priors and sampler settings except that we ran three Markov chain Monte Carlo chains of 10,000 iterations each, including 5,000 warm-up iterations. See Supplemental Method, “Bayesian Outcome Noninferiority Analysis,” for implementation details and Table S3 for diagnostics. All models showed satisfactory convergence and no divergent transitions.

Effect-size benchmarks commonly characterize correlations or standardized regression coefficients with an absolute magnitude of approximately .10 as small (Gignac & Szodorai, 2016; Nieminen, 2022), and we therefore treated $\beta = -.10$ as the smallest negative association considered substantively meaningful. Because the analysis asked whether higher AI-reference EDGE was associated with meaningfully poorer outcomes, we used a one-sided noninferiority criterion rather than testing two-sided equivalence (Lakens et al., 2018). Following an HDI-based posterior decision rule (Kruschke, 2018), we concluded that an association was noninferior when the lower bound of the 95% HDI exceeded −.10. For each outcome, we report the posterior median, 95% HDI, and $P(\beta > -.10)$. A noninferior result therefore indicates that higher AI-reference EDGE was not associated with a meaningfully poorer outcome.

### *EDGE and Assessments of Originality and Creativity*

The human ratings and automatic scores used in this study were designed to evaluate how original or creative a response was judged to be. EDGE, by contrast, quantifies how weakly a response is represented within a specified comparison distribution. It therefore describes a response's distributional semantic position rather than directly evaluating its originality or creativity. This distinction is especially important in relation to AI. A response may be judged highly original even when similar responses are common among AI outputs. Conversely, a response may occupy a semantic region rarely reached by AI without being judged especially original. The relationship between EDGE and these assessments is therefore empirical rather than definitional.

Drawing on the semantic-disjunction hypothesis (Moon et al., 2026), we expected this relationship to differ between human and AI responses. Among human responses, creativity should increase as EDGE scores increase (i.e., responses occupy semantic positions more weakly represented in the shared response space). Among AI responses, however, high originality or creativity scores need not correspond to weak semantic representation, as automatic assessments may assign high scores to responses located in regions densely occupied by AI outputs. We therefore expected a positive relationship between EDGE and originality or creativity among human responses, but a weaker relationship among AI responses. A systematically weaker or reversed relationship among AI responses would be inconsistent with treating EDGE as interchangeable with assessments designed to evaluate originality or creativity. We tested this prediction using the joint human-AI resampling draws from the Coverage Analysis. Because the comparison required source-comparable evaluations, we included only automatic assessments produced by the same evaluator for human and AI responses. Within each resampling draw, we estimated an EDGE × source interaction to quantify the difference between human and AI responses in the relationship between EDGE and the task-relevant outcomes. Across draws, we summarized the task-level human coefficient, AI coefficient, and their paired difference using medians and the shortest empirical intervals containing 95% of the resampled estimates (95% resampling intervals). See Supplemental Method, "EDGE and Assessments of Originality and Creativity," for model specifications.

## Results

### Coverage Analysis

Across all three tasks, human responses fell, on average, in semantically sparser regions of the pooled human–AI response corpus than matched AI responses. We quantified this coverage difference within each resampling draw as mean human EDGE minus mean AI EDGE. Figure 3 shows the resampling distributions of these human-minus-AI contrasts. Median contrasts were 0.991 (95% resampling interval [0.931, 1.048]) for admissions essays, 1.295 (95% resampling interval [1.126, 1.426]) for AUT, and 1.141 (95% resampling interval [.993, 1.278]) for CPS. Median contrasts were also positive in all eight strata in the admissions essay sample (range = 0.419–1.655) and in both AUT objects (brick = 1.226; knife = 1.375; see Table S2).

**Figure 3**

*Distributions of Human–AI Differences in Mean EDGE Across Tasks*

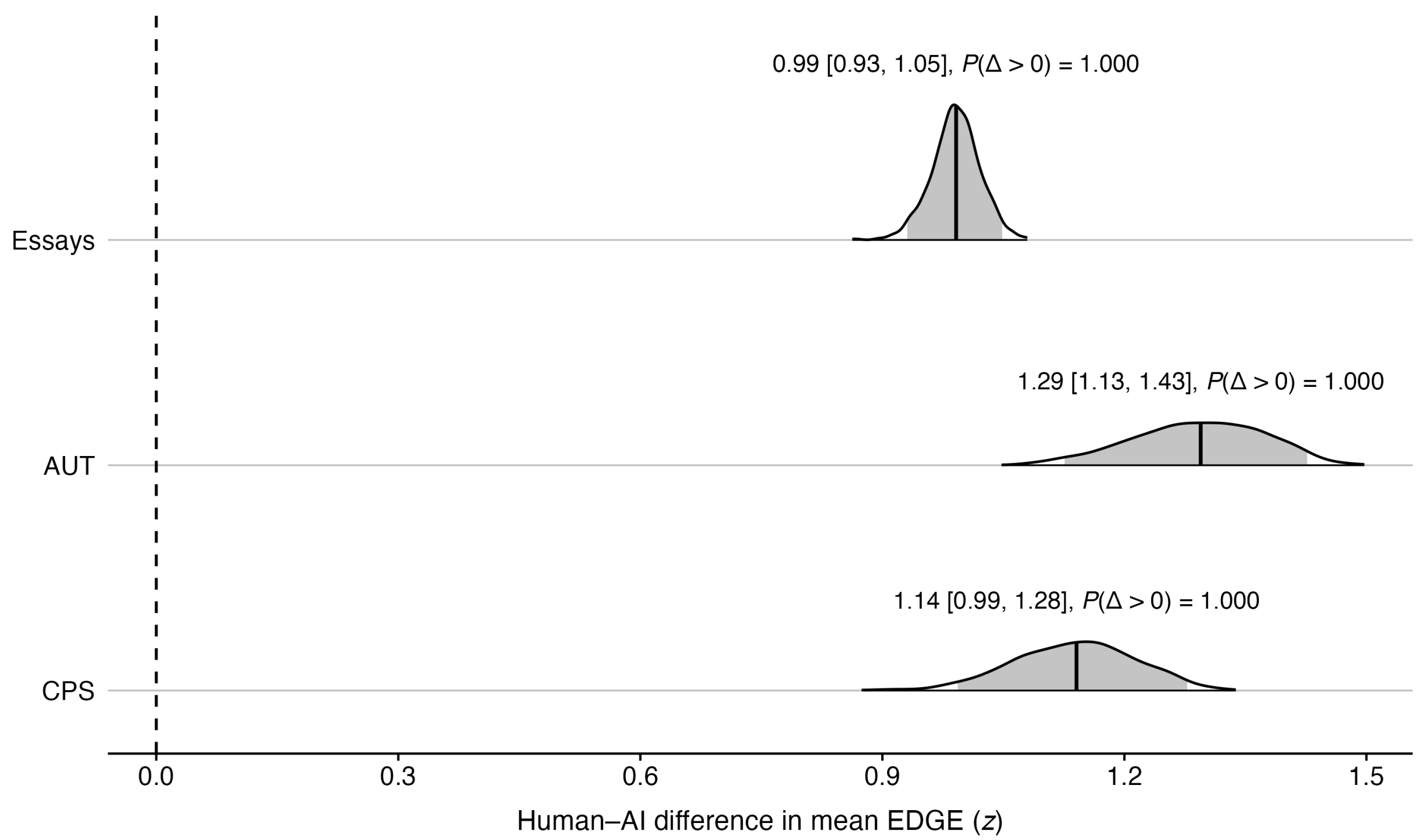

*Note.* Ridges show the empirical resampling distributions of differences in mean EDGE between human and matched AI response groups, calculated as human minus AI. Vertical lines indicate task-specific medians. Labels report the median, 95% resampling interval, and $P(\Delta > 0)$, the proportion of 1,000 resampled comparisons in which mean human EDGE exceeded mean AI EDGE ($\Delta$ = human − AI). The dashed vertical line at zero indicates no

difference between the human and matched AI response groups. AUT = Alternative Uses Task; CPS = Creative Problem-Solving Task.

**Outcome Noninferiority Analysis**

We examined nine outcomes: essay creativity, SAT, and college GPA for admissions essays; human-rated and automatically assessed originality for AUT; and human-rated originality and quality plus automatically assessed originality and effectiveness for CPS. Across all nine outcomes, higher AI-reference EDGE was not associated with meaningfully poorer performance: Every 95% HDI had a lower bound above the −.10 noninferiority margin (Figure 4; Table S4). Four associations were credibly positive beyond noninferiority, whereas the remaining five were compatible with no association but not with a meaningfully negative one.

The clearest positive associations emerged in admissions essays. AI-reference EDGE was most strongly associated with essay creativity as evaluated by an expert-tuned model (median $\beta = .35$, 95% HDI [.31, .40], $P(\beta > -.10) = 1.000$) and, more modestly, with SAT (median $\beta = .11$, 95% HDI [.04, .18], $P(\beta > -.10) = 1.000$). Its association with college GPA was centered near zero but remained above the noninferiority margin (median $\beta = .03$, 95% HDI [−.02, .08], $P(\beta > -.10) = 1.000$).

For AUT, the 95% HDIs included zero for both human-rated originality (median $\beta = .03$, 95% HDI [−.06, .12], $P(\beta > -.10) = .997$) and CLAUS originality (median $\beta = .01$, 95% HDI [−.09, .10], $P(\beta > -.10) = .987$), but their lower bounds remained above the noninferiority margin.

For CPS, AI-reference EDGE was positively associated with both human-rated originality (median $\beta = .20$, 95% HDI [.13, .26], $P(\beta > -.10) = 1.000$) and CPS-AI originality (median $\beta = .17$, 95% HDI [.11, .23], $P(\beta > -.10) = 1.000$). Its associations with human-rated quality (median β = .02, 95% HDI [−.05, .08], $P(\beta > -.10) > .999$) and CPS-AI effectiveness (median β = .04, 95% HDI [−.02, .11], $P(\beta > -.10) > .999$) were centered near zero but remained within the noninferiority range.

**Figure 4**

*Posterior Distributions of the EDGE–Outcome Association*

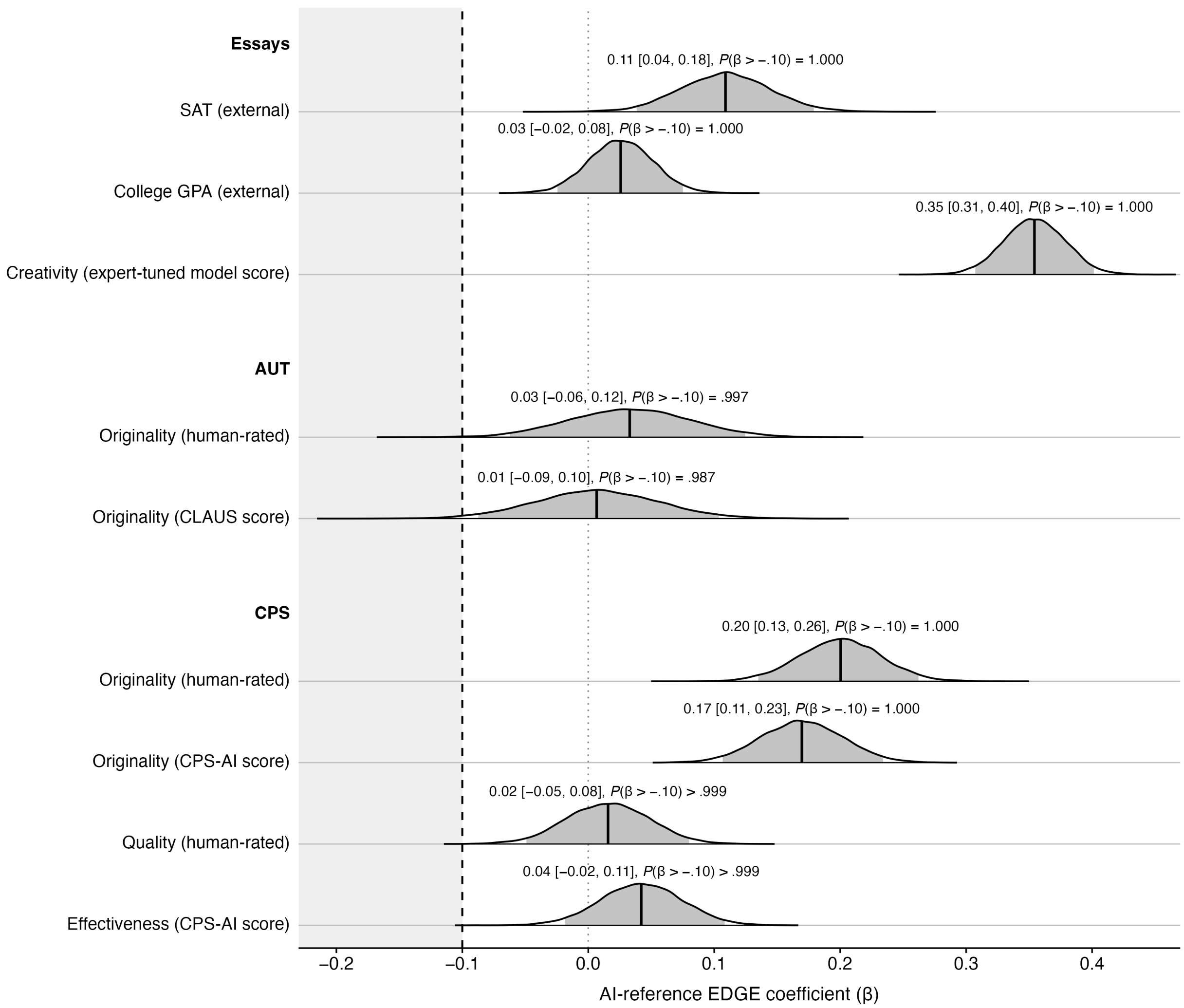

*Note.* Ridges show the posterior distributions of the standardized AI-reference EDGE coefficients. The shaded portion of each ridge indicates the 95% highest density interval (HDI), thin horizontal lines show the range of retained posterior draws, and vertical lines indicate posterior medians. Labels report the posterior median, 95% HDI,

and $P(\beta > -.10)$. The dashed line marks the prespecified noninferiority margin of −.10; the light-gray background indicates values at or below this margin, and the dotted line marks zero. Parenthetical labels identify the assessment source or scorer. AUT = Alternative Uses Task; CPS = Creative Problem-Solving Task; CLAUS = automatic originality scorer for the AUT; CPS-AI = automatic scorer for the CPS.

### EDGE and Assessments of Originality and Creativity

Across all three tasks, higher EDGE was consistently associated with higher originality or creativity scores among human responses (Figure 5; Table S5). Median standardized slopes were .67 (95% resampling interval [.60, .75]) for essay creativity, .45 (95% resampling interval [.29, .61]) for AUT CLAUS originality, and .54 (95% resampling interval [.25, .80]) for CPS-AI originality. By contrast, the corresponding association among AI responses was small and positive for essay creativity, median $\beta = .05$ (95% resampling interval [.02, .08]); negative for AUT CLAUS originality, median $\beta = -.11$ (95% resampling interval [−.18, −.02]); and positive but substantially smaller than the human slope for CPS-AI originality, median $\beta = .11$ (95% resampling interval [.01, .22]). Median human–AI differences in the standardized EDGE coefficients were $\Delta\beta = .62$ (95% resampling interval [.55, .71]) for essays, $\Delta\beta = .56$ (95% resampling interval [.37, .75]) for AUT, and $\Delta\beta = .43$ (95% resampling interval [.14, .74]) for CPS. Using an absolute standardized coefficient of .10 as a small-effect benchmark, the 95% resampling intervals for the standardized EDGE coefficients among human responses and for the human–AI differences in these coefficients lay entirely above .10.

## Figure 5

*EDGE–Assessment Associations Within Human and AI Responses*

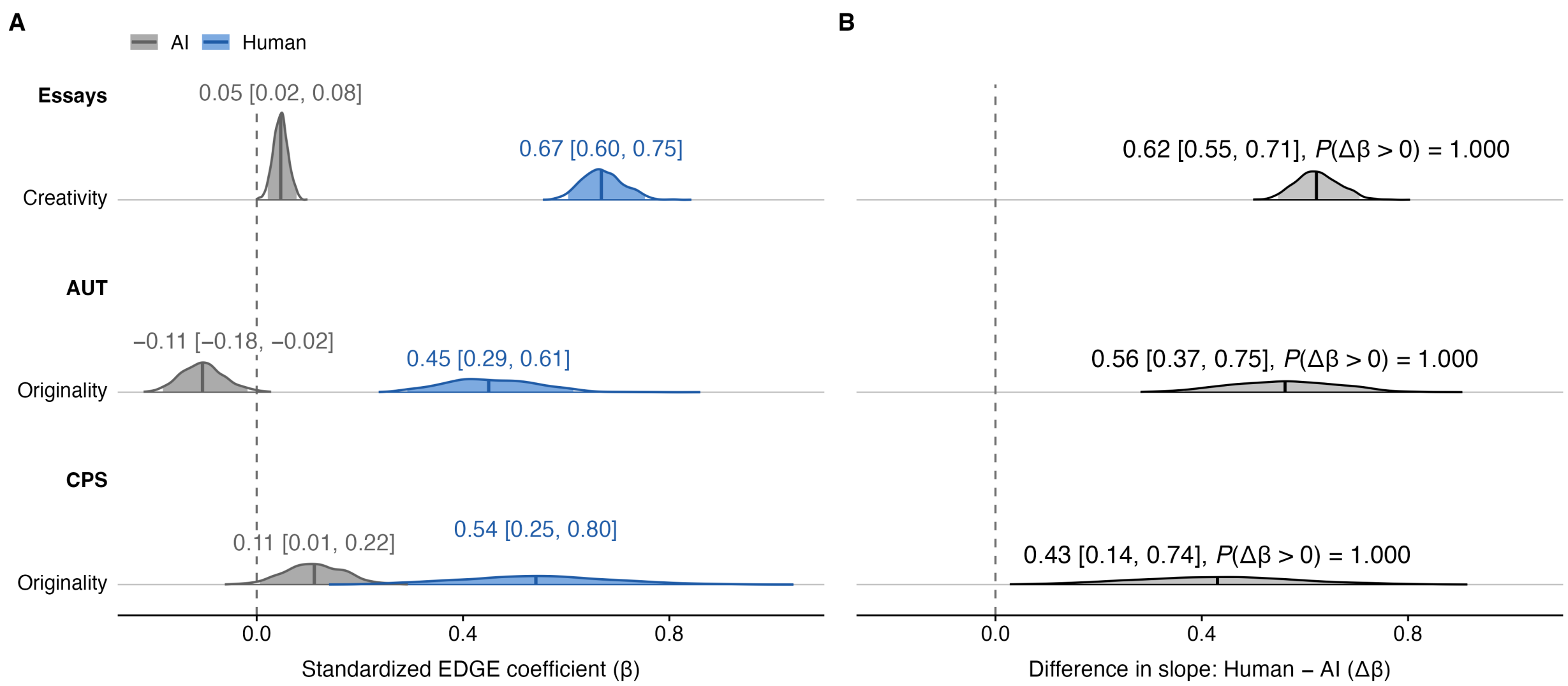

*Note.* Ridges show the empirical resampling distributions of standardized EDGE coefficients. Panel A shows the coefficients separately for human and AI responses; Panel B shows their differences, calculated as human minus AI. Vertical lines indicate medians, shaded regions indicate 95% resampling intervals, and thin horizontal lines show the realized ranges. Labels report medians and 95% resampling intervals. Panel B additionally reports $P(\Delta\beta > 0)$, the proportion of 1,000 resampled comparisons in which the human coefficient exceeded the AI coefficient. The dashed lines in Panels A and B mark zero. Within each resampling draw and matched response group, EDGE and assessment scores were standardized across the combined human and AI responses. Positive values in Panel B indicate stronger EDGE–assessment associations among human responses. AUT = Alternative Uses Task; CPS = Creative Problem-Solving Task; CLAUS = automatic originality scorer for AUT; CPS-AI = automatic originality scorer for CPS. See Figure S1 in the Supplemental Materials for descriptive scatterplots of response-level mean EDGE scores and automatic assessment scores.

## Discussion

This study tested whether human ideas that diverge from AI homogenization reflect valuable thinking. Across large samples of admissions essays from applicants to three institutions and two independent creativity tasks, a computational method developed to identify ideas that expand a semantic idea space (EDGE) showed that human responses consistently reached regions sparsely covered by matched AI outputs (AI semantic blind spots). Human ideas in those blind spots were as good as or better than human ideas more similar to AI with respect to task-relevant outcomes. The findings were consistent across diverse forms of idea generation, indicating that the idea diversity that distinguishes humans from AI may be advantageous.

EDGE scores sensitively distinguished human and AI content on the basis of heterogeneity across all datasets, tasks, and prompts. In every essay sampling stratum, for both Alternative Uses Task (AUT) objects, and for the Creative Problem-Solving Task (CPS) scenario, human responses occupied, on average, sparser regions of the shared semantic space than matched AI responses. To assess whether this heterogeneity reflected valuable thinking, we examined associations between AI-reference EDGE scores and task-relevant evaluations. In essays, ideas that fell further outside the space of AI-generated responses–those "outside the bots"–received higher creativity evaluations. In CPS, these ideas received higher originality ratings from both humans and AI evaluators. Associations with originality in the AUT and with quality and effectiveness in CPS also met the noninferiority criterion, although none was credibly positive. These findings suggest that AI's semantic blind spots encompass valuable differences in human thinking, rather than simply responses poorly suited to the task.

Extending EDGE analysis from ideas to their authors, we found evidence that ideas in AI's semantic blind spots may reflect their authors' academic strengths. Specifically, college applicants whose essays extended further into these blind spots scored higher on the SAT. The association with college GPA was nominally positive and met the noninferiority criterion, although it was not credibly positive. The significance of human heterogeneity therefore extends beyond evaluations of the writing itself: differences less well represented by AI were also associated with academic capability in the people who generated those ideas. That this connection emerged in the high-stakes context of college admissions reinforces the ecological validity and real-world implications of the findings. Human ideas that depart from typical AI responses may express meaningful strengths of their producers even in settings where the way people present themselves can have substantial consequences.

Human ideas that diverge from AI may thus confer advantages at two levels. At the individual level, occupying an AI semantic blind spot was associated with better and not worse outcomes. At the population level, a broader range of good ideas raises the collective likelihood of advances. Creative breakthroughs in science, industry, the arts, and elsewhere arise from variation followed by selection (Campbell, 1960), and the number of high-impact contributions rises with the total number of contributions generated (Simonton, 1997). Innovation likewise proceeds by generating many candidate ideas in parallel and selecting few for development (Kornish & Ulrich, 2011), and the quality of the best idea in a pool depends on the number and variability of the ideas considered (Girotra et al., 2010). More variable search produces more breakthroughs (Fleming, 2001), atypical combinations of prior work predict the highest-impact science (Uzzi et al., 2013), and groups of diverse problem solvers can outperform groups of uniformly high-ability ones (Hong & Page, 2004). More ideas improve the odds of progress when the added ideas are good, and the present results indicate that the human ideas underrepresented by AI meet that condition. The contraction of collective idea diversity documented here and across a substantial extant body of research (Anderson et al., 2024; Doshi & Hauser, 2024; Meincke et al., 2025; Moon et al., 2025, 2026; Padmakumar & He, 2024; Sourati et al., 2026; Wenger & Kenett, 2025) would therefore prune the pool of candidate solutions without a compensating gain in quality, a collective cost that is not revealed by ratings of individual responses (Burton et al., 2024).

Concerns that AI threatens innovation have arisen because AI-generated ideas often appear creative but actually narrow the idea space (Doshi & Hauser, 2024; Meincke et al., 2025; Moon et al., 2026). EDGE demonstrated the ability to detect such ideas. Although EDGE scores were strongly and positively associated with originality and creativity assessments among human responses in all three tasks, the corresponding association among AI responses was small for essays and the CPS and reversed in direction for the AUT: Higher originality ratings did not consistently correspond to greater semantic sparsity among AI responses across tasks. In contrast, the most creative human ideas were those that expanded the idea space. This dissociation is consistent with recent findings of AI "semantic disjunction" (Moon et al., 2026), showing that essays produced after the widespread availability of AI, and essays known to be generated by and with AI, used more varied words but expressed a narrower range of underlying ideas. Human and AI raters judged the AI-influenced text as more creative because of its varied wording. The association between word-level and idea-level diversity was positive in human writing but weakened or reversed in GPT writing. Here, the disjunction between creativity ratings and semantic distinctness was evident with EDGE, which considers not only differences between individual responses but also the density and collective structure of ideas. These findings and those of Moon et al. (2026) caution against treating high creativity ratings of AI outputs as evidence that those outputs broaden a collective idea space. Evaluative assessments and EDGE capture different properties of a response, and both appear necessary for understanding what AI adds to, and subtracts from, the range of available ideas.

To assess the value of human idea heterogeneity, we compared ideas produced entirely by humans against those generated by AI. This approach (i.e., only using human-authored essays from 2022 or earlier) minimized ambiguity about the source of the ideas being compared. We anticipate that the capacity of EDGE to identify ideas that expand the space will have applicability to questions concerning not only human-only and AI-only idea spaces, but also human–AI hybrid creativity. Comparisons of competing AI models and progressive model versions, as well as measurement of efforts to mitigate tradeoffs in AI creativity and diversity, may be fruitful directions. In general, increasing temperature and other parameter-based levers to increase AI heterogeneity (e.g., *top_p*) have been found to result in decreased coherence, so AI ideas that expand idea space the most may be of lower value (Peeperkorn et al., 2024; Schaeffer et al., 2025), although prompting- and training-based approaches may fare better (Ismayilzada et al., 2025; Lanchantin et al., 2025; Meincke et al., 2024). EDGE offers a response-level measure for testing whether such interventions extend a model's semantic reach without sacrificing value.

Taken together, our findings suggest that human ideas beyond AI's homogenized clusters are not simply inferior ideas that AI can safely omit. Across three tasks, humans produced valuable ideas in regions that AI rarely reached. This has potentially broad implications for innovation because progress depends not only on selecting the best ideas from an existing set, but also on generating ideas beyond that set. In our data, humans—not AI—expanded the available idea space. If AI increasingly shapes the ideas people produce, the pool of ideas available to science, business, and society may grow only when people think beyond what AI generates.

## Acknowledgments

The authors used generative AI tools to support code development and figure preparation and to improve the clarity and readability of the manuscript. All resulting materials were reviewed and revised as needed by the authors, who take full responsibility for the final content.

## Data and Code Availability

The data and code supporting the findings of this study will be made publicly available upon publication.

## References

ACT, Inc. & College Board. (2018). *Guide to the 2018 ACT/SAT concordance*. ACT, Inc. and College Board. https://www.act.org/content/dam/act/unsecured/documents/ACT-SAT-Concordance-Information.pdf

Alvero, A. J., Lee, J., Regla-Vargas, A., Kizilcec, R. F., Joachims, T., & Antonio, A. L. (2024). Large language models, social demography, and hegemony: Comparing authorship in human and synthetic text. *Journal of Big Data*, *11*(1), 138. https://doi.org/10.1186/s40537-024-00986-7

Anderson, B. R., Shah, J. H., & Kreminski, M. (2024). Homogenization effects of large language models on human creative ideation. *Proceedings of the 16th Conference on Creativity & Cognition*, 413–425. https://doi.org/10.1145/3635636.3656204

Beaty, R. E., Kenett, Y. N., Christensen, A. P., Rosenberg, M. D., Benedek, M., Chen, Q., Fink, A., Qiu, J., Kwapil, T. R., Kane, M. J., & Silvia, P. J. (2018). Robust prediction of individual creative ability from brain functional connectivity. *Proceedings of the National Academy of Sciences*, *115*(5), 1087–1092. https://doi.org/10.1073/pnas.1713532115

Bender, E. M., Gebru, T., McMillan-Major, A., & Shmitchell, S. (2021). On the Dangers of Stochastic Parrots: Can Language Models Be Too Big? 🦜. *Proceedings of the 2021 ACM Conference on Fairness, Accountability, and Transparency*, 610–623. https://doi.org/10.1145/3442188.3445922

Bezdek, J. C., Ehrlich, R., & Full, W. (1984). FCM: The fuzzy c-means clustering algorithm. *Computers & Geosciences*, *10*(2–3), 191–203. https://doi.org/10.1016/0098-3004(84)90020-7

Bürkner, P.-C. (2017). brms: An R package for Bayesian multilevel models using Stan. *Journal of Statistical Software*, *80*(1), 1–28. https://doi.org/10.18637/jss.v080.i01

Burton, J. W., Lopez-Lopez, E., Hechtlinger, S., Rahwan, Z., Aeschbach, S., Bakker, M. A., Becker, J. A., Berditchevskaia, A., Berger, J., Brinkmann, L., Flek, L., Herzog, S. M., Huang, S., Kapoor, S., Narayanan, A., Nussberger, A.-M., Yasseri, T., Nickl, P., Almaatouq, A., … Hertwig, R. (2024). How large language models can reshape collective intelligence. *Nature Human Behaviour*, *8*(9), 1643–1655. https://doi.org/10.1038/s41562-024-01959-9

Campbell, D. T. (1960). Blind variation and selective retention in creative thought as in other knowledge processes. *Psychological Review*, *67*(6), 380–400. https://doi.org/10.1037/h0040373

Campello, R. J. G. B., Moulavi, D., Zimek, A., & Sander, J. (2015). Hierarchical Density Estimates for Data Clustering, Visualization, and Outlier Detection. *ACM Transactions on Knowledge Discovery from Data*, *10*(1), 1–51. https://doi.org/10.1145/2733381

Doshi, A. R., & Hauser, O. P. (2024). Generative AI enhances individual creativity but reduces the collective diversity of novel content. *Science Advances*, *10*(28), eadn5290. https://doi.org/10.1126/sciadv.adn5290

Fleming, L. (2001). Recombinant uncertainty in technological search. *Management Science*, *47*(1), 117–132. https://doi.org/10.1287/mnsc.47.1.117.10671

Fortier, A., Chen, H., & West, P. (2026). *Is Convergence Inevitable? Tracing Output Homogeneity Back to Base Models* (arXiv:2608.11426). arXiv. https://doi.org/10.48550/arXiv.2608.11426

Gignac, G. E., & Szodorai, E. T. (2016). Effect size guidelines for individual differences researchers. *Personality and Individual Differences*, *102*, 74–78. https://doi.org/10.1016/j.paid.2016.06.069

Girotra, K., Terwiesch, C., & Ulrich, K. T. (2010). Idea generation and the quality of the best idea. *Management Science*, *56*(4), 591–605. https://doi.org/10.1287/mnsc.1090.1144

Holzner, N., Maier, S., & Feuerriegel, S. (2025). *Generative AI and creativity: A systematic literature review and meta-analysis* (arXiv:2505.17241). arXiv. https://doi.org/10.48550/arXiv.2505.17241

Hong, L., & Page, S. E. (2004). Groups of diverse problem solvers can outperform groups of high-ability problem solvers. *Proceedings of the National Academy of Sciences*, *101*(46), 16385–16389. https://doi.org/10.1073/pnas.0403723101

Ismayilzada, M., Laverghetta, A., Jr., Luchini, S. A., Patel, R., Bosselut, A., van der Plas, L., & Beaty, R. E. (2025). Creative preference optimization. *Findings of the Association for Computational Linguistics: EMNLP 2025*, 9580–9609. https://doi.org/10.18653/v1/2025.findings-emnlp.509

Jiang, L., Chai, Y., Li, M., Liu, M., Fok, R., Dziri, N., Tsvetkov, Y., Sap, M., Albalak, A., & Choi, Y. (2025). Artificial hivemind: The open-ended homogeneity of language models (and beyond). In D. Belgrave, C. Zhang, H. Lin, R. Pascanu, P. Koniusz, M. Ghassemi, & N. Chen (Eds.), *Advances in neural information processing systems* (Vol. 38). Curran Associates, Inc. https://proceedings.neurips.cc/paper_files/paper/2025/file/754d5a526a5ee5a47220664a0eb92751-Paper-Datasets_and_Benchmarks_Track.pdf

Kirk, R., Mediratta, I., Nalmpantis, C., Luketina, J., Hambro, E., Grefenstette, E., & Raileanu, R. (2024). *Understanding the Effects of RLHF on LLM Generalisation and Diversity*. arXiv. https://doi.org/10.48550/arXiv.2310.06452

Klypa, R., & Cherednichenko, O. (2026). *Diversity in Large Language Models under Supervised Fine-Tuning* (arXiv:2605.00195). arXiv. https://doi.org/10.48550/arXiv.2605.00195

Kornish, L. J., & Ulrich, K. T. (2011). Opportunity spaces in innovation: Empirical analysis of large samples of ideas. *Management Science*, *57*(1), 107–128. https://doi.org/10.1287/mnsc.1100.1247

Kruschke, J. K. (2018). Rejecting or Accepting Parameter Values in Bayesian Estimation. *Advances in Methods and Practices in Psychological Science*, *1*(2), 270–280. https://doi.org/10.1177/2515245918771304

Lakens, D., Scheel, A. M., & Isager, P. M. (2018). Equivalence Testing for Psychological Research: A Tutorial. *Advances in Methods and Practices in Psychological Science*, *1*(2), 259–269. https://doi.org/10.1177/2515245918770963

Lanchantin, J., Chen, A., Dhuliawala, S., Yu, P., Weston, J., Sukhbaatar, S., & Kulikov, I. (2025). Diverse preference optimization. *arXiv Preprint arXiv:2501.18101*. https://arxiv.org/abs/2501.18101

Luchini, S. A., Maliakkal, N. T., DiStefano, P. V., Laverghetta, A., Patterson, J. D., Beaty, R. E., & Reiter-Palmon, R. (2025). Automated scoring of creative problem solving with large language models: A comparison of originality and quality ratings. *Psychology of Aesthetics, Creativity, and the Arts*. https://doi.org/10.1037/aca0000736

Luchini, S. A., Pronchick, J., Ceh, S. M., Kaufman, J. C., Johnson, D. R., Rafner, J., & Beaty, R. (2025). *The Roles of Idea Generation and Elaboration in Human-AI Collaborative Creativity*. PsyArXiv. https://doi.org/10.31234/osf.io/xm2f5_v3

Meincke, L., Mollick, E. R., & Terwiesch, C. (2024). *Prompting diverse ideas: Increasing AI idea variance*. SSRN. https://doi.org/10.2139/ssrn.4708466

Meincke, L., Nave, G., & Terwiesch, C. (2025). ChatGPT decreases idea diversity in brainstorming. *Nature Human Behaviour*, *9*(6), 1107–1109. https://doi.org/10.1038/s41562-025-02173-x

Moon, K., Green, A. E., & Kushlev, K. (2025). Homogenizing effect of large language models (LLMs) on creative diversity: An empirical comparison of human and ChatGPT writing. *Computers in Human Behavior: Artificial Humans*, *6*, 100207. https://doi.org/10.1016/j.chbah.2025.100207

Moon, K., Kushlev, K., Bank, A., Luttges, B. L., Viskontas, I., Kaufman, J. C., Johnson, D. R., Duckworth, A. L., & Green, A. (2026). *The link between diverse words and original ideas is weakening in the AI-era college admissions*. PsyArXiv. https://doi.org/10.31234/osf.io/jsz58_v8

Murthy, S. K., Ullman, T., & Hu, J. (2025). One fish, two fish, but not the whole sea: Alignment reduces language models' conceptual diversity. *Proceedings of the 2025 Conference of the Nations of the Americas Chapter of the Association for Computational Linguistics: Human Language Technologies (Volume 1: Long Papers)*, 11241–11258. https://doi.org/10.18653/v1/2025.naacl-long.561

Nieminen, P. (2022). Application of standardized regression coefficient in meta-analysis. *BioMedInformatics*, *2*(3), 434–458. https://doi.org/10.3390/biomedinformatics2030028

Padmakumar, V., & He, H. (2024). Does writing with language models reduce content diversity? *International Conference on Learning Representations 2024 (ICLR 2024)*. https://proceedings.iclr.cc/paper_files/paper/2024/hash/02dec8877fb7c6aa9a79f81661baca7c-Abstract-Conference.html

Patel, N., Crossman, J., Aggarwal, E., & Wenger, E. (2026). *Are LLMs becoming similarly creative? Evidence from three years of models* (arXiv:2608.19437). arXiv. https://doi.org/10.48550/arXiv.2608.19437

Patterson, J. D., Merseal, H. M., Johnson, D. R., Agnoli, S., Baas, M., Baker, B. S., Barbot, B., Benedek, M., Borhani, K., Chen, Q., Christensen, J. F., Corazza, G. E., Forthmann, B., Karwowski, M., Kazemian, N., Kreisberg-Nitzav, A., Kenett, Y. N., Link, A., Lubart, T., ... Beaty, R. E. (2023). Multilingual semantic distance: Automatic verbal creativity assessment in many languages. *Psychology of Aesthetics, Creativity, and the Arts*, *17*(4), 495–507. https://doi.org/10.1037/aca0000618

Patterson, J. D., Pronchick, J., Panchanadikar, R., Fuge, M., van Hell, J., Miller, S. R., Johnson, D. R., & Beaty, R. E. (2025). CAP: The creativity assessment platform for online testing and automated scoring. *Behavior Research Methods*, *57*, Article 264. https://doi.org/10.3758/s13428-025-02761-9

Peeperkorn, M., Kouwenhoven, T., Brown, D., & Jordanous, A. (2024). *Is Temperature the Creativity Parameter of Large Language Models?* (arXiv:2405.00492). arXiv. https://doi.org/10.48550/arXiv.2405.00492

Reimers, N., & Gurevych, I. (2019). Sentence-BERT: Sentence Embeddings using Siamese BERT-Networks. *Proceedings of the 2019 Conference on Empirical Methods in Natural*

*Language Processing and the 9th International Joint Conference on Natural Language Processing (EMNLP-IJCNLP)*, 3980–3990. https://doi.org/10.18653/v1/D19-1410

Reiter-Palmon, R., Mumford, M. D., & Threlfall, K. V. (1998). Solving Everyday Problems Creatively: The Role of Problem Construction and Personality Type. *Creativity Research Journal*, *11*(3), 187–197. https://doi.org/10.1207/s15326934crj1103_1

Resnik, P. (2025). Large Language Models Are Biased Because They Are Large Language Models. *Computational Linguistics*, *51*(3), 885–906. https://doi.org/10.1162/coli_a_00558

Schaeffer, R., Kazdan, J., & Denisov-Blanch, Y. (2025). Turning Down the Heat: A Critical Analysis of Min-p Sampling in Language Models. *arXiv Preprint arXiv:2506.13681*. https://arxiv.org/abs/2506.13681

Simonton, D. K. (1997). Creative productivity: A predictive and explanatory model of career trajectories and landmarks. *Psychological Review*, *104*(1), 66–89. https://doi.org/10.1037/0033-295X.104.1.66

Solatorio, A. V. (2024a). *GISTEmbed: Guided In-sample Selection of Training Negatives for Text Embedding Fine-tuning* (arXiv:2402.16829). arXiv. https://doi.org/10.48550/arXiv.2402.16829

Solatorio, A. V. (2024b). *NoInstruct-small-Embedding-v0* [Computer software]. Hugging Face. https://huggingface.co/avsolatorio/NoInstruct-small-Embedding-v0

Song, K., Tan, X., Qin, T., Lu, J., & Liu, T.-Y. (2020). *MPNet: Masked and Permuted Pre-training for Language Understanding* (Version 2). arXiv. https://doi.org/10.48550/ARXIV.2004.09297

Sourati, Z., Ziabari, A. S., & Dehghani, M. (2026). The homogenizing effect of large language models on human expression and thought. *Trends in Cognitive Sciences*, S1364661326000033. https://doi.org/10.1016/j.tics.2026.01.003

Uzzi, B., Mukherjee, S., Stringer, M., & Jones, B. (2013). Atypical combinations and scientific impact. *Science*, *342*(6157), 468–472. https://doi.org/10.1126/science.1240474

Wenger, E., & Kenett, Y. (2025). *We're different, we're the same: Creative homogeneity across LLMs* (arXiv:2501.19361). arXiv. https://doi.org/10.48550/arXiv.2501.19361

**Supplemental Materials for**

**Thinking Outside the Bots:**

**Human Ideas in AI's Semantic Blind Spots Are Not Inferior**

## Supplemental Method

### Datasets and Response Sources

#### *Admissions Essays*

**Human Responses and Matching.** We analyzed 1,600 admissions essays written by college applicants from three U.S. institutions: an elite private university, a regional private university, and a selective liberal-arts college. We retained eight strata containing at least 200 essays from the 2018–2022 matriculated sample: one at the elite private university, four at the regional private university, and three at the selective liberal-arts college. We randomly sampled 200 essays without replacement from each stratum, yielding a human sample of 1,600 essays that was balanced across strata. Three prompts were shared by the regional private university and the selective liberal-arts college, so the eight essay sampling strata represented five distinct prompts. Matriculation was operationalized by the availability of a finite cumulative college GPA. For prompts shared across institutions, the same prompt-specific AI pool supplied the corresponding institution–prompt comparisons; human essays remained grouped by institution and prompt.

**AI Response Generation.** AI essays were generated using GPT-5-mini (gpt-5-mini-2025-08-07). The AI sample comprised 200 essays for each of the five prompts, yielding 1,000 essays in total. Target lengths were based on the median word count of human essays for each prompt: 639 words for personal growth, 616 for an engaging topic, 637 for a challenge or failure, 646 for the personal or creative essay, and 639 for background or identity. The generation code capped output at three times the requested word count in tokens and otherwise left sampling settings at their API defaults.

**Generation Instructions.** The generation instruction was: "You are an excellent college admissions essay writer. Write a compelling, personal essay of about {target_words} words. Be creative, authentic, reflective, and specific." The user message supplied the corresponding essay prompt, preceded by "Essay prompt:".

**Admissions Essay Prompts.** The five admissions essay prompts were as follows.

*Personal or creative essay: As [institution name] is a diverse community, the Admissions Committee would like to know more about you in your own words. Please submit a brief essay, either personal or creative, which you feel best describes you.*

*Personal growth: Discuss an accomplishment, event, or realization that sparked a period of personal growth and a new understanding of yourself or others.*

*Engaging topic: Describe a topic, idea, or concept you find so engaging that it makes you lose all track of time. Why does it captivate you? What or who do you turn to when you want to learn more?*

*Challenge or failure: The lessons we take from obstacles we encounter can be fundamental to later success. Recount a time when you faced a challenge, setback, or failure. How did it affect you, and what did you learn from the experience?*

*Background or identity: Some students have a background, identity, interest, or talent that is so meaningful they believe their application would be incomplete without it. If this sounds like you, then please share your story.*

The institution name in the personal or creative essay prompt is replaced here with a bracketed label. The generation request used the institution name in the original prompt.

### *Alternative Uses Task*

**Human Responses and Matching.** We analyzed Alternative Uses Task (AUT) responses from Patterson et al. (2023) for two object prompts: brick and knife. The matched AI response pool comprised 481 GPT-5-mini pseudo-participant bundles generated for the same brick and knife prompts, with one AI bundle corresponding to each human bundle and containing the same number of individual uses. The retained 3,425 human uses formed the 481 bundles analyzed here. Individual uses remained together within their human or AI participant–object bundle throughout response sampling and analysis.

**AI Response Generation.** AI response bundles were generated using GPT-5-mini (gpt-5-mini-2025-08-07). The AI sample comprised 481 bundles, with one bundle corresponding to each human participant–object bundle and containing the same number of individual uses. Each AI bundle was generated in a separate request through the Responses API, without overriding temperature or other sampling settings. Returned responses were parsed as an array, leading and trailing whitespace was removed, and internal whitespace was normalized. An output was accepted only when the array contained the requested number of nonempty uses and no case-insensitive duplicate uses.

**Generation Instructions.** For each request, {response_count} was the number of uses in the matched human bundle, and {object_name} was either brick or knife. The instruction was: "You are completing an Alternative Uses Task. Return only JSON with a key named responses whose value is an array of strings. The array must contain exactly {response_count} distinct, concise alternate uses. Avoid numbering, explanations, markdown, or ordinary/default uses."

*The user prompt was: "Object: {object_name}*

*Task: Give exactly {response_count} creative alternate uses for this object."*

***Creative Problem-Solving Task***

**Human Responses and Matching.** For the Creative Problem-Solving Task (CPS), we used the Becky's Problem responses from the CPS data reported by Luchini et al. (2025). Becky's Problem focuses on an interpersonal dilemma with competing practical and relational concerns (Reiter-Palmon et al., 1998). The human sample consisted of 1,018 undergraduate-written responses across the Becky datasets, and the AI reference set consisted of 1,018 GPT-5-mini responses generated for the same Becky's Problem prompt. Each participant contributed one complete response, and each complete human or AI solution was treated as one response-level unit.

**AI Response Generation.** AI solutions were generated using GPT-5-mini (gpt-5-mini-2025-08-07). The AI sample comprised 1,018 responses to Becky's Problem, generated in separate requests submitted through the Batch API on October 10, 2026. To approximate the mean length of the human responses (53.77 words and 2.58 sentences), each request asked for a single creative solution of approximately 54 words in about three sentences. Reasoning effort was set to medium, and max_completion_tokens was set to 8,192, including reasoning tokens. Temperature, top_p, and seed were not specified. The instruction targeted the human mean length rather than the variability of human response lengths.

**Generation Instructions.** The generation instructions were as follows.

*System instruction: Follow all instructions carefully. Follow all steps of the step by step instructions. This is for academic research purposes only.*

*User instruction: Your task is generate a single creative solution to the following prompt: Becky is a college student who works part-time at Mark's Pizzeria. Mark, the owner of the restaurant, has treated Becky very well. He gave her a job that she needs to help pay her rent when no other business would employ her because she was arrested for shoplifting three years ago. Mark also lets Becky work around her school schedule, and has asked if she wants to be a shift manager in the summers. Becky's roommate Jim also works at the pizzeria, but Jim has been causing a lot of problems at work. He always avoids doing his job, treats customers rudely, and makes a lot of mistakes with orders. Jim recently began stealing food from the pizzeria. Two days ago the pizzeria was short-staffed, so Jim and Becky were the only employees left at closing time. Jim made 10 extra pizzas and took them home to a party he was hosting without paying for them. Becky feels like she needs to do something about Jim's behavior. However, Becky is hesitant to tell Mark about Jim because Jim is a good friend to Becky. Becky also needs Jim to have a job so he can pay his portion of their rent. Becky does not know what to do.*
*Aim for approximately 54 words in about 3 sentences.*
*Don't output any additional conversational text.  Do not label the single creative solution with any text, number, or any label at all.*
*It is very important you follow all instructions!*

Table S1a summarizes sample composition. Table S1b reports response length means and standard deviations.

**Table S1a**

*Sample Composition and Matched Response Groups*

| Task | Human sample or unit | Prompt or object | Human *n* | AI pool *n* |
|---|---|---|---|---|
| Essays | Elite private | Personal or creative essay | 200 | 200 |
| Essays | Regional private | Personal growth | 200 | 200 |
| Essays | Regional private | Engaging topic | 200 | 200 |
| Essays | Regional private | Challenge or failure | 200 | 200 |
| Essays | Regional private | Background or identity | 200 | 200 |
| Essays | Selective liberal-arts | Personal growth | 200 | 200 |
| Essays | Selective liberal-arts | Challenge or failure | 200 | 200 |
| Essays | Selective liberal-arts | Background or identity | 200 | 200 |
| AUT | Participant–object bundles | Brick | 240 | 240 |
| AUT | Participant–object bundles | Knife | 241 | 241 |
| CPS | Complete responses | Becky's Problem | 1,018 | 1,018 |

*Note.* Admissions essays comprised 1,600 human essays and 1,000 unique AI essays. Three prompts were shared across two institutions, and the corresponding AI pools were reused across institution–prompt comparisons; AI pool counts in those rows must not be summed as unique essays. AUT comprised 481 human and 481 AI bundles, each source containing 3,425 individual uses. The 481 human bundles came from 241 participants. CPS comprised 1,018 human and 1,018 AI responses. Each Coverage draw used 120 human and 120 AI response units within a matched group. AUT = Alternative Uses Task; CPS = Creative Problem-Solving Task.

**Table S1b**

*Response Length by Matched Response Group*

| Sample | Prompt or object | Human Words M (SD) | AI Words M (SD) | Human Sentences/ responses M (SD) | AI Sentences/ responses M (SD) |
|---|---|---|---|---|---|
| Elite private | Personal or creative essay | 641.70 (89.24) | 662.52 (66.54) | 32.38 (7.48) | 31.79 (4.28) |
| Regional private | Personal growth | 609.75 (118.85) | 687.11 (109.93) | 31.45 (7.89) | 39.00 (6.76) |
| Regional private | Engaging topic | 577.16 (148.21) | 566.11 (148.44) | 28.65 (10.25) | 27.04 (7.54) |
| Regional private | Challenge or failure | 594.81 (134.22) | 654.80 (99.10) | 32.02 (10.32) | 39.30 (6.49) |
| Regional private | Background or identity | 617.62 (134.53) | 639.63 (126.97) | 32.50 (10.36) | 32.92 (7.09) |
| Selective liberal-arts | Personal growth | 619.46 (50.64) | 687.11 (109.93) | 32.27 (7.40) | 39.00 (6.76) |
| Selective liberal-arts | Challenge or failure | 619.06 (54.32) | 654.80 (99.10) | 34.38 (8.49) | 39.30 (6.49) |
| Selective liberal-arts | Background or identity | 618.39 (56.45) | 639.63 (126.97) | 32.66 (6.98) | 32.92 (7.09) |
| AUT | Brick | 27.78 (16.02) | 79.50 (30.29) | 7.25 (3.08) | 7.25 (3.08) |
| AUT | Knife | 25.65 (14.61) | 66.63 (21.62) | 7.00 (3.33) | 7.00 (3.33) |
| CPS | Becky’s Problem | 53.77 (37.94) | 53.94 (1.23) | 2.58 (1.74) | 3.00 (0.04) |

*Note*. Values are means (sample standard deviations) per essay, AUT participant–object bundle, or complete CPS response, using the full pools listed in Table S1a. The sentence/response columns report sentence counts for admissions essays and CPS, and response counts for AUT. For AUT, each response is an individual use (idea), and counts indicate the number of ideas within each participant–object bundle. AUT word counts are summed across those ideas. AUT = Alternative Uses Task; CPS = Creative Problem-Solving Task.

## Outcome Measures

### *External Person-Level Outcomes*

External person-level outcomes for admissions essay authors included SAT and cumulative college grade point average (GPA), obtained from university records. ACT composite scores were converted to SAT equivalents using the official 2018 ACT/SAT concordance (ACT, Inc. & College Board, 2018). When both SAT and ACT scores were available, we retained the higher SAT-scale score. The SAT analysis excluded the regional private university because SAT availability was sparse in that sample. It included 794 essay authors with available SAT-scale scores from the elite private university and selective liberal-arts college. Cumulative GPA was available for all 1,600 essay authors.

### *Human-Rated Response Outcomes*

Human-rated response outcomes were available only for human-authored AUT and CPS responses. For the AUT, we summarized originality at the participant–object level by averaging the use-level scores (Patterson et al., 2023). CPS originality and quality were complete-solution ratings (Luchini et al., 2025). Human-rated originality was available for all 481 AUT bundles. CPS human-rated originality and quality were available for 1,017 of the 1,018 human responses. These outcome-specific counts were used in the corresponding models; the full response pool remained available for the Coverage Analysis.

### *Automatic Assessments of Creativity, Originality, and Effectiveness*

To enable direct comparisons between response sources, we applied the same automatic assessment within each task to both human and matched AI-generated responses. Admissions essays were scored using a GPT-4.1-mini evaluator fine-tuned to predict creativity judgments provided by 22 creativity researchers (Moon et al., 2026). AUT and CPS responses were scored using Penn State's Creativity Assessment Platform (CAP; Patterson et al., 2025). For the AUT, Cross-Lingual Alternate Uses Scoring (CLAUS) assessed the originality of each AUT use, and we averaged these scores within each participant–object bundle. For the CPS, we used CPS-AI to evaluate the originality and effectiveness of each complete response (Luchini et al., 2025).

The same evaluator scored all 1,600 human admissions essays and all 1,000 matched AI essays. Each essay was submitted as a complete text, and the evaluator returned a single numeric creativity score on a 0–1 scale. For the AUT, CLAUS evaluated all 6,850 individual uses across the human and AI pools. CLAUS received the object name and each individual use and returned an originality score on a 0–1 scale, and the resulting use-level scores were averaged within bundles to obtain 962 bundle-level scores. For CPS, CPS-AI evaluated all 2,036 complete human and AI responses. The evaluator received the full Becky's Problem scenario and each complete solution and returned originality and effectiveness predictions. CPS-AI analyses used the raw predictions without clipping. Although the documented nominal scale ranged from −2 to 2, 22 human originality predictions and 28 human effectiveness predictions exceeded the upper bound. Two human responses exceeded the upper bound on both outcomes, yielding 48 distinct human responses with out-of-range predictions. No predictions for AI responses fell outside the nominal range.

## EDGE Computation

### *Text Representation*

To implement the text representation procedure described in the main text, we used avsolatorio/NoInstruct-small-Embedding-v0 (Solatorio, 2024b), sentence-transformers/all-

mpnet-base-v2 (Reimers & Gurevych, 2019; Song et al., 2020), and avsolatorio/GIST-Embedding-v0 (Solatorio, 2024a). Before sentence segmentation, leading and trailing whitespace was removed and consecutive whitespace characters were collapsed. Admissions essays and CPS responses were segmented into sentences before embedding. Each model used its corresponding tokenizer, with padding within batches and truncation at 512 tokens. Representations were extracted from the first token of the final hidden layer with the model in evaluation mode. These vectors entered clustering directly, without additional length normalization or dimensionality reduction.

***Clustering Parameter Selection***

Clustering parameters were selected independently for each of the three embedding models and reselected whenever the scoring corpus changed, using outcome-blind criteria. For HDBSCAN, minPts controls the neighborhood size used to estimate local density. We evaluated minPts = {3, 5, 8, 11, 14, 17} and selected the value minimizing mean GLOSH across embedded rows. For fuzzy c-means, k specifies the number of clusters, and m controls the fuzziness of cluster membership, with larger values allowing responses to have more evenly distributed membership across clusters. We evaluated all 12 combinations of k = {3, 4, 5, 6, 7, 8} and m = {1.05, 1.10}, selecting the combination maximizing the range of cluster-membership ambiguity. Neither human–AI coverage contrasts nor outcome scores informed selection. Both algorithms used Euclidean distances. Each fuzzy c-means parameter combination was run once from a randomized starting configuration. The algorithm repeatedly updated cluster centers and membership values until convergence or a maximum of 100 iterations. Candidates were evaluated in ascending order of minPts or, for fuzzy c-means, k followed by m; exact ties were resolved in favor of the first candidate.

### *Aggregation and Standardization*

EDGE components were computed across the resulting task-specific embedding rows and then averaged within each analytic unit: across sentences for each admissions essay and CPS response, and across uses for each AUT participant–object bundle. The two components were computed separately in each of the three embedding spaces, producing six component scores for every analytic unit. Each embedded row contributed equally to its unit's component mean. Units could contribute different numbers of sentences or uses to the clustering corpus; AUT bundles contained 1–23 uses (M = 7.1).

Each of the six unit-level components was standardized within the analysis-specific scoring set. The final EDGE score was the arithmetic mean of these six standardized components. For the Coverage Analysis, standardization used all 240 human and AI units within a matched response group and draw. For AI-reference scoring, component standardization used the human target units sharing a prompt, object, or problem and reference panel. Before outcome modeling, the resulting EDGE scores were standardized again within each institution–prompt sampling stratum, AUT object, or CPS problem, separately for each panel.

## Bayesian Outcome Noninferiority Analysis

### *AI-Reference EDGE*

For each human response and AI reference panel, we combined the human response's embedded sentences or uses with those of the AI reference responses. Both clustering methods and their parameter searches were applied to this combined corpus. Only the human response's component scores were retained and averaged across its sentences or uses. Other human responses did not enter this calculation. Thus, human responses shared matched AI reference panels but were scored through separate clustering calculations. Different reference panels could contain some of the same AI responses.

### *Model Specification*

A separate model was fitted for each continuous outcome using the default Gaussian family in brms. To account for uncertainty arising from AI reference-panel composition, we explicitly supplied the standard error of each response's mean EDGE across panels as its measurement-error standard deviation. The observed mean EDGE was modeled as normally distributed around a latent EDGE value. The latent-EDGE submodel estimated residual variation among responses, distinct from measurement error, following the standard Gaussian specification. We explicitly fixed residual correlations between the latent-EDGE and outcome submodels to zero, so that no additional residual association was estimated beyond the modeled EDGE–outcome relationship.

Outcomes were standardized within matched response groups using the observations eligible for each outcome model. Missing outcome values were not imputed. Task-level coefficients averaged slopes across eight institution–prompt strata for essay creativity and GPA, four eligible strata for SAT, two objects for the AUT, and the single CPS problem. Model convergence and sampling diagnostics are reported in Table S3.

## EDGE and Assessments of Originality and Creativity

Within each matched response group and draw, we regressed the standardized automatic assessment on standardized EDGE, response source, and their interaction. Source was coded as 0 for AI responses and 1 for human responses. The EDGE coefficient therefore represented the AI slope, and adding the interaction coefficient yielded the human slope. The interaction coefficient represented the human-minus-AI slope difference.

Task-level human slopes, AI slopes, and their differences were calculated within each draw by equally averaging the corresponding estimates across the eight admissions essay strata or the two AUT objects. CPS involved a single problem and required no further aggregation. The number of positive interactions across draws summarized the directional consistency of the source difference.

**Visual Explanation of the Two EDGE Components**

The visualization used 75 human and 75 AI admissions essays, 240 human and 240 AI AUT Brick bundles, and 300 human and 300 GPT-5-mini CPS responses. Two-dimensional coordinates were obtained using PaCMAP (Wang et al., 2021), with human and AI responses sharing a display space within each task. To make the underlying concepts visually clear, display coordinates were rescaled, with additional systematic transformations of radial spread and proximity to cluster centers for AUT and CPS to clarify the spatial patterns used to illustrate the two components. Cluster-membership ambiguity illustrated responses that did not align strongly with a single semantic cluster. The outlyingness component captured by GLOSH was illustrated through sparsity in the display space, highlighting responses located away from dense concentrations of ideas. These illustrative outlyingness values were converted to percentile ranks across the pooled human and AI points within each task, with shared color scales supporting comparison between sources. The visualization provided a conceptual illustration of the two EDGE components; the reported analyses calculated these components in the original high-dimensional embedding spaces.

## Supplemental Results

### Coverage Analysis

Table S2 reports the human-minus-AI contrasts separately for all eight admissions essay strata, both AUT objects, and the CPS scenario, together with the task-level aggregates.

**Table S2**

*Human–AI Differences in Mean EDGE by Matched Response Group*

| Sample | Prompt or aggregation | Median | 95% resampling interval | Positive draws |
|---|---|---|---|---|
| Elite private | Personal or creative essay | 1.655 | [1.521, 1.708] | 1,000 |
| Regional private | Personal growth | 0.419 | [0.219, 0.625] | 1,000 |
| Regional private | Engaging topic | 1.080 | [0.962, 1.199] | 1,000 |
| Regional private | Challenge or failure | 0.443 | [0.251, 0.645] | 998 |
| Regional private | Background or identity | 1.034 | [0.876, 1.191] | 1,000 |
| Selective liberal-arts | Personal growth | 0.988 | [0.826, 1.131] | 1,000 |
| Selective liberal-arts | Challenge or failure | 1.040 | [0.882, 1.222] | 1,000 |
| Selective liberal-arts | Background or identity | 1.294 | [1.193, 1.428] | 1,000 |
| AUT | Brick | 1.226 | [1.064, 1.355] | 1,000 |
| AUT | Knife | 1.375 | [1.068, 1.582] | 1,000 |
| CPS | Becky's Problem | 1.141 | [0.993, 1.278] | 1,000 |
| Admissions essays, stratum-equal | Task aggregate | 0.991 | [0.931, 1.048] | 1,000 |
| AUT, object-equal | Task aggregate | 1.295 | [1.126, 1.426] | 1,000 |
| CPS | Task aggregate | 1.141 | [0.993, 1.278] | 1,000 |
| Admissions essays, institution-equal | Sensitivity | 1.165 | [1.106, 1.214] | 1,000 |

*Note.* Contrasts are mean human EDGE minus mean AI EDGE. Intervals are the shortest empirical intervals containing 95% of 1,000 draw-level contrasts. Positive draws are counts out of 1,000, not *p*-values or posterior probabilities. Primary task aggregates equally weight eight admissions strata, two AUT objects, or the single CPS scenario. The admissions sensitivity first averages within institution and then equally weights the three institutions. EDGE = expansion of divergent generative expression. AUT = Alternative Uses Task; CPS = Creative Problem-Solving Task.

All 11 matched response groups had positive medians and 95% resampling intervals entirely above zero. The contrast was positive in 1,000 of 1,000 draws for ten groups and in 998 of 1,000 draws for the regional private university's challenge-or-failure prompt.

As a sensitivity analysis, we first averaged contrasts across prompts within each institution and then weighted the three institutions equally. The institution-equal admissions contrast was 1.165 (95% resampling interval [1.106, 1.214]), positive in 1,000 of 1,000 draws. Thus, the direction of the admissions coverage result was unchanged when institutions, rather than institution–prompt strata, received equal weight.

## Outcome Noninferiority Analysis

All models showed satisfactory convergence and no divergent transitions. Across the nine models, the largest fit-wide R-hat was 1.00133; the smallest bulk and tail effective sample sizes were 3,101.79 and 5,306.25, respectively. No model reached the maximum tree depth. Table S3 reports diagnostics for each model.

**Table S3**

*Bayesian Model Convergence and Sampling Diagnostics*

| Task | Outcome | Max $\hat{R}$ | Min bulk ESS | Min tail ESS | Divergences | Tree-depth hits | Min E-BFMI |
|---|---|---|---|---|---|---|---|
| Essays | SAT | 1.00082 | 4,425 | 5,306 | 0 | 0 | 0.993 |
| Essays | Essay creativity | 1.00133 | 5,013 | 5,589 | 0 | 0 | 0.981 |
| Essays | Cumulative GPA | 1.00082 | 5,530 | 7,103 | 0 | 0 | 1.021 |
| AUT | Human-rated originality | 1.00124 | 3,314 | 6,376 | 0 | 0 | 0.737 |
| AUT | CLAUS originality | 1.00133 | 3,102 | 5,437 | 0 | 0 | 0.725 |
| CPS | Human-rated originality | 1.00066 | 4,620 | 6,517 | 0 | 0 | 0.993 |
| CPS | CPS-AI originality | 1.00082 | 4,781 | 6,652 | 0 | 0 | 0.994 |
| CPS | Human-rated quality | 1.00039 | 5,119 | 7,005 | 0 | 0 | 0.997 |
| CPS | CPS-AI effectiveness | 1.00060 | 4,797 | 6,183 | 0 | 0 | 0.977 |

*Note.* All models used three chains of 10,000 iterations, including 5,000 warm-up iterations per chain, yielding 15,000 retained posterior draws per outcome. $\hat{R}$ and effective sample sizes (ESS) summarize the full fitted model; minima and maxima are taken across model parameters as indicated. Tree-depth hits count transitions reaching the sampler's maximum tree depth. E-BFMI = energy Bayesian fraction of missing information; the minimum is across chains. AUT = Alternative Uses Task; CPS = Creative Problem-Solving Task. GPA = grade point average. CLAUS = Cross-Lingual Alternate Uses Scoring; CPS-AI = automatic scorer for Creative Problem Solving.

Every 95% HDI had a lower bound above the −.10 noninferiority margin (Table S4).

**Table S4**

*Posterior Associations of AI-Reference EDGE with Task-Relevant Outcomes*

| Task | Outcome | *n* | Median β | 95% HDI | *P*(β > −0.10) | Conclusion |
|---|---|---|---|---|---|---|
| Essays | SAT | 794 | 0.109 | [0.039, 0.179] | 1.000 | Positive |
| Essays | Essay creativity (expert-tuned model) | 1,600 | 0.354 | [0.307, 0.401] | 1.000 | Positive |
| Essays | College GPA | 1,600 | 0.026 | [−0.024, 0.075] | 1.000 | Noninferior |
| AUT | Human-rated originality | 481 | 0.033 | [−0.062, 0.125] | .997 | Noninferior |
| AUT | CLAUS originality | 481 | 0.007 | [−0.087, 0.104] | .987 | Noninferior |
| CPS | Human-rated originality | 1,017 | 0.200 | [0.135, 0.262] | 1.000 | Positive |
| CPS | CPS-AI originality | 1,018 | 0.170 | [0.107, 0.234] | 1.000 | Positive |
| CPS | Human-rated quality | 1,017 | 0.016 | [−0.049, 0.080] | > .999 | Noninferior |
| CPS | CPS-AI effectiveness | 1,018 | 0.042 | [−0.018, 0.108] | > .999 | Noninferior |

*Note.* β values are posterior medians, and all coefficients are standardized. Noninferiority was supported when the lower bound of the 95% HDI exceeded −0.10. "Positive" indicates that the entire 95% HDI was above zero; "Noninferior" indicates that the 95% HDI included zero but excluded values at or below −0.10. Essay creativity was evaluated by an expert-tuned model. Estimates are reported to three decimals. HDI = highest density interval; GPA = grade point average. EDGE = expansion of divergent generative expression. AUT = Alternative Uses Task; CPS = Creative Problem-Solving Task. CLAUS = Cross-Lingual Alternate Uses Scoring; CPS-AI = automatic scorer for Creative Problem Solving.

### EDGE and Assessments of Originality and Creativity

Table S5 reports the human and AI coefficients and their paired differences. Each task-level human–AI difference was positive in 1,000 of 1,000 draws.

**Table S5**

*Associations Between EDGE and Originality or Creativity Assessments Among Human and AI Responses*

| Task | Assessment | Human β [95% interval] | AI β [95% interval] | Human–AI Δβ [95% interval] |
|---|---|---|---|---|
| Essays | Essay creativity | 0.668 [0.604, 0.753] | 0.047 [0.022, 0.077] | 0.622 [0.548, 0.706] |
| AUT | CLAUS originality | 0.450 [0.292, 0.614] | −0.105 [−0.182, −0.018] | 0.561 [0.367, 0.752] |
| CPS | CPS-AI originality | 0.541 [0.247, 0.800] | 0.112 [0.006, 0.224] | 0.430 [0.135, 0.743] |

*Note.* Entries are medians with 95% resampling intervals from the distributions of task-level standardized coefficients across resampling iterations. Within each iteration, the corresponding human and AI coefficient estimates were paired, and Human–AI Δβ was calculated as $\beta_{Human} - \beta_{AI}$. The reported Δβ and its 95% resampling interval therefore summarize the distribution of these paired differences, rather than differences between the reported medians or interval bounds. Positive Δβ values indicate a stronger EDGE–assessment association among human responses. Estimates are reported to three decimals. EDGE = expansion of divergent generative expression. AUT = Alternative Uses Task; CPS = Creative Problem-Solving Task. CLAUS = Cross-Lingual Alternate Uses Scoring; CPS-AI = automatic scorer for Creative Problem Solving.

**Stability Across AI-Reference Panels**

We evaluated the stability of AI-reference EDGE across the 10 reference panels using two-way consistency intraclass correlation coefficients (ICCs). Scores were standardized separately within each panel and matched response group, as in the Outcome Noninferiority Analysis. We calculated ICC(C,1) for a single panel and ICC(C,10) for the mean of 10 panels, with F-based 95% confidence intervals. Analyses were conducted separately for each institution-prompt stratum, AUT object, and CPS problem. Consistency was the relevant criterion because the analysis concerns relative differences among human responses after within-panel standardization.

Across the eight admissions essay strata, single-panel ICCs ranged from .945 to .987 and mean-of-10-panel ICCs ranged from .994 to .999. For AUT, the corresponding ICCs were .797 and .975 for brick and .793 and .975 for knife. For CPS, they were .909 and .990, respectively (Table S6). Thus, averaging across the 10 panels yielded stable relative EDGE scores under the current reference-sampling procedure.

**Table S6**

*Consistency of AI-Reference EDGE Across Reference Panels*

| Sample | Prompt or object | *n* | ICC(C,1) [95% CI] | ICC(C,10) [95% CI] |
|---|---|---|---|---|
| Elite private | Personal or creative essay | 200 | .987 [.984, .990] | .999 [.998, .999] |
| Regional private | Personal growth | 200 | .963 [.956, .970] | .996 [.995, .997] |
| Regional private | Engaging topic | 200 | .945 [.934, .955] | .994 [.993, .995] |
| Regional private | Challenge or failure | 200 | .958 [.949, .966] | .996 [.995, .996] |
| Regional private | Background or identity | 200 | .982 [.978, .985] | .998 [.998, .999] |
| Selective liberal-arts | Personal growth | 200 | .965 [.957, .972] | .996 [.996, .997] |
| Selective liberal-arts | Challenge or failure | 200 | .957 [.948, .965] | .996 [.995, .996] |
| Selective liberal-arts | Background or identity | 200 | .987 [.984, .990] | .999 [.998, .999] |
| AUT | Brick | 240 | .797 [.765, .827] | .975 [.970, .980] |
| AUT | Knife | 241 | .793 [.760, .824] | .975 [.969, .979] |
| CPS | Becky’s Problem | 1,018 | .909 [.901, .916] | .990 [.989, .991] |

*Note*. *ICC*(C,1) estimates the consistency reliability of EDGE scores computed using a single reference panel; *ICC*(C,10) estimates the consistency reliability of EDGE scores averaged across 10 reference panels. Both coefficients are estimated using scores from all 10 panels. Each panel contained 120 matched AI responses. EDGE was standardized separately within each panel and matched response group. Confidence intervals are model-based F intervals. The *n* column counts essays, participant–object bundles, or complete CPS responses.

## Descriptive Associations

Figure S1 displays descriptive associations between EDGE and four automatic assessments: essay creativity, CLAUS originality, CPS-AI originality, and CPS-AI effectiveness.

## Figure S1

*Descriptive Associations Between EDGE and Originality, Creativity, or Effectiveness Assessments Among Human and AI Responses*

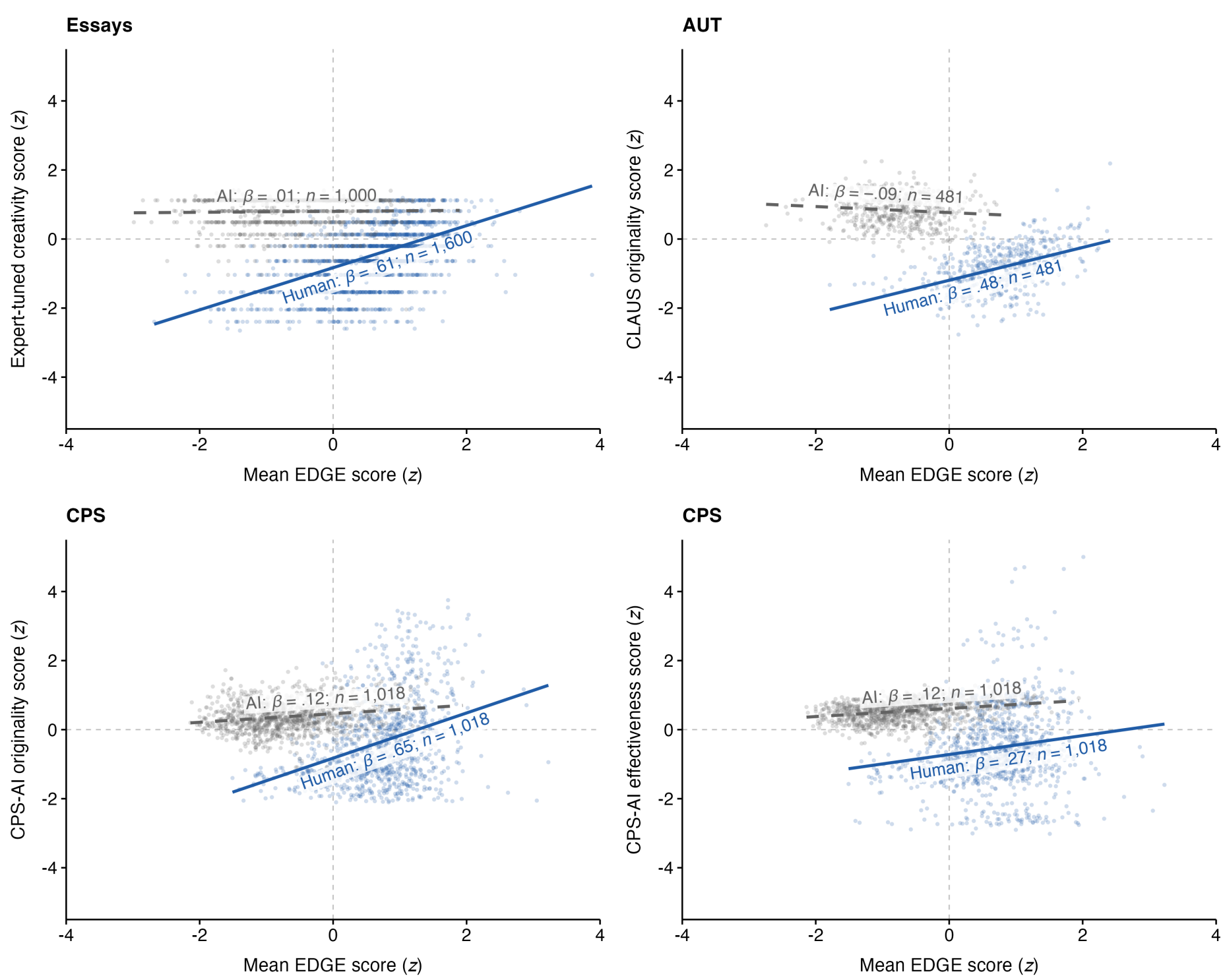

*Note.* Each point represents one human or AI response unit. EDGE scores were standardized across human and AI responses within each matched group and draw, then averaged for each response across the draws in which it appeared. Assessment scores were standardized across human and AI responses within each panel. Regression lines were fitted separately for human and AI responses using these response-level mean EDGE scores; labels report the resulting slopes (β) and sample sizes. AUT = Alternative Uses Task; CPS = Creative Problem-Solving Task; CLAUS = Cross-Lingual Alternate Uses Scoring; CPS-AI = automatic scorer for creative problem solving.

## Supplemental References

ACT, Inc. & College Board. (2018). *Guide to the 2018 ACT/SAT concordance*. ACT, Inc. and College Board. https://www.act.org/content/dam/act/unsecured/documents/ACT-SAT-Concordance-Information.pdf

Luchini, S. A., Maliakkal, N. T., DiStefano, P. V., Laverghetta, A., Patterson, J. D., Beaty, R. E., & Reiter-Palmon, R. (2025). Automated scoring of creative problem solving with large language models: A comparison of originality and quality ratings. *Psychology of Aesthetics, Creativity, and the Arts*. https://doi.org/10.1037/aca0000736

Moon, K., Kushlev, K., Bank, A., Luttges, B. L., Viskontas, I., Kaufman, J. C., Johnson, D. R., Duckworth, A. L., & Green, A. (2026). *The link between diverse words and original ideas is weakening in the AI-era college admissions*. PsyArXiv. https://doi.org/10.31234/osf.io/jsz58_v8

Patterson, J. D., Merseal, H. M., Johnson, D. R., Agnoli, S., Baas, M., Baker, B. S., Barbot, B., Benedek, M., Borhani, K., Chen, Q., Christensen, J. F., Corazza, G. E., Forthmann, B., Karwowski, M., Kazemian, N., Kreisberg-Nitzav, A., Kenett, Y. N., Link, A., Lubart, T., ... Beaty, R. E. (2023). Multilingual semantic distance: Automatic verbal creativity assessment in many languages. *Psychology of Aesthetics, Creativity, and the Arts*, *17*(4), 495–507. https://doi.org/10.1037/aca0000618

Patterson, J. D., Pronchick, J., Panchanadikar, R., Fuge, M., van Hell, J., Miller, S. R., Johnson, D. R., & Beaty, R. E. (2025). CAP: The creativity assessment platform for online testing and automated scoring. *Behavior Research Methods*, *57*, Article 264. https://doi.org/10.3758/s13428-025-02761-9

Reimers, N., & Gurevych, I. (2019). Sentence-BERT: Sentence Embeddings using Siamese BERT-Networks. *Proceedings of the 2019 Conference on Empirical Methods in Natural Language Processing and the 9th International Joint Conference on Natural Language Processing (EMNLP-IJCNLP)*, 3980–3990. https://doi.org/10.18653/v1/D19-1410

Reiter-Palmon, R., Mumford, M. D., & Threlfall, K. V. (1998). Solving Everyday Problems Creatively: The Role of Problem Construction and Personality Type. *Creativity Research Journal*, *11*(3), 187–197. https://doi.org/10.1207/s15326934crj1103_1

Solatorio, A. V. (2024a). *GISTEmbed: Guided In-sample Selection of Training Negatives for Text Embedding Fine-tuning* (arXiv:2402.16829). arXiv. https://doi.org/10.48550/arXiv.2402.16829

Solatorio, A. V. (2024b). *NoInstruct-small-Embedding-v0* [Computer software]. Hugging Face. https://huggingface.co/avsolatorio/NoInstruct-small-Embedding-v0

Song, K., Tan, X., Qin, T., Lu, J., & Liu, T.-Y. (2020). *MPNet: Masked and Permuted Pre-training for Language Understanding* (Version 2). arXiv. https://doi.org/10.48550/ARXIV.2004.09297

Wang, Y., Huang, H., Rudin, C., & Shaposhnik, Y. (2021). Understanding How Dimension Reduction Tools Work: An Empirical Approach to Deciphering t-SNE, UMAP, TriMap, and PaCMAP for Data Visualization. *Journal of Machine Learning Research*, *22*(201), 1–73.